\documentclass[sigconf]{acmart}
\acmSubmissionID{897}

\makeatletter
\DeclareRobustCommand\onedot{\futurelet\@let@token\@onedot}
\def\@onedot{\ifx\@let@token.\else.\null\fi\xspace}

\makeatother

\newcommand{\ceil}[1]{\lceil {#1} \rceil}
\newcommand{\ignore}[1]{}   

\newcommand{\mpage}[2]
{
\begin{minipage}{#1\linewidth}\centering
#2
\end{minipage}
}

\usepackage{enumitem}

\AtBeginDocument{%
  \providecommand\BibTeX{{%
    \normalfont B\kern-0.5em{\scshape i\kern-0.25em b}\kern-0.8em\TeX}}}

\copyrightyear{2024}
\acmYear{2024}
\copyrightyear{2024}
\acmYear{2024}
\setcopyright{rightsretained}
\acmConference[SIGGRAPH Conference Papers '24]{Special Interest Group on Computer Graphics and Interactive Techniques Conference Conference Papers '24}{July 27-August 1, 2024}{Denver, CO, USA}
\acmBooktitle{Special Interest Group on Computer Graphics and Interactive Techniques Conference Conference Papers '24 (SIGGRAPH Conference Papers '24), July 27-August 1, 2024, Denver, CO, USA}
\acmDOI{10.1145/3641519.3657488}
\acmISBN{979-8-4007-0525-0/24/07}

\usepackage{float}
\begin{document}

\title{Modeling Ambient Scene Dynamics for Free-view Synthesis}


\author{Meng-Li Shih}
\affiliation{%
  \institution{University of Washington}
  \city{Seattle}
  \state{Washington}
  \country{USA}
  }
\email{mlshih@cs.washington.edu}

\author{Jia-Bin Huang}
\affiliation{%
  \institution{University of Maryland}
  \city{College Park}
  \state{Maryland}
  \country{USA}
  }
\email{jbhuang@umd.edu}

\author{Changil Kim}
\affiliation{%
  \institution{Meta}
  \city{Seattle}
  \state{Washington}
  \country{USA}
  }
\email{changil@meta.com}

\author{Rajvi Shah}
\affiliation{%
  \institution{Meta}
  \city{Seattle}
  \state{Washington}
  \country{USA}
  }
\email{rajvishah@meta.com}

\author{Johannes Kopf}
\affiliation{%
  \institution{Meta}
  \city{Seattle}
  \state{Washington}
  \country{USA}
  }
\email{jkopf@meta.com}

\author{Chen Gao}
\affiliation{%
  \institution{Meta}
  \city{Seattle}
  \state{Washington}
  \country{USA}
  }
\email{gaochen@meta.com}


\begin{abstract}
We introduce a novel method for dynamic free-view synthesis of an ambient scenes from a monocular capture bringing a immersive quality to the viewing experience. 
Our method builds upon the recent advancements in 3D Gaussian Splatting (3DGS) that can faithfully reconstruct complex static scenes. 
Previous attempts to extend 3DGS to represent dynamics have been confined to bounded scenes or require multi-camera captures, and often fail to generalize to unseen motions, limiting their practical application. 
Our approach overcomes these constraints by leveraging the periodicity of ambient motions to learn the motion trajectory model, coupled with careful regularization. 
We also propose important practical strategies to improve the visual quality of the baseline 3DGS static reconstructions and to improve memory efficiency critical for GPU-memory intensive learning. 
We demonstrate high-quality photorealistic novel view synthesis of several ambient natural scenes with intricate textures and fine structural elements.
We show that our method significantly outperforms prior methods both qualitatively and quantitatively. 
Project page: \textcolor{magenta}{\url{https://ambientGaussian.github.io/}}
\end{abstract} 


\begin{teaserfigure}
  \centering
  \includegraphics[width=0.85\textwidth]{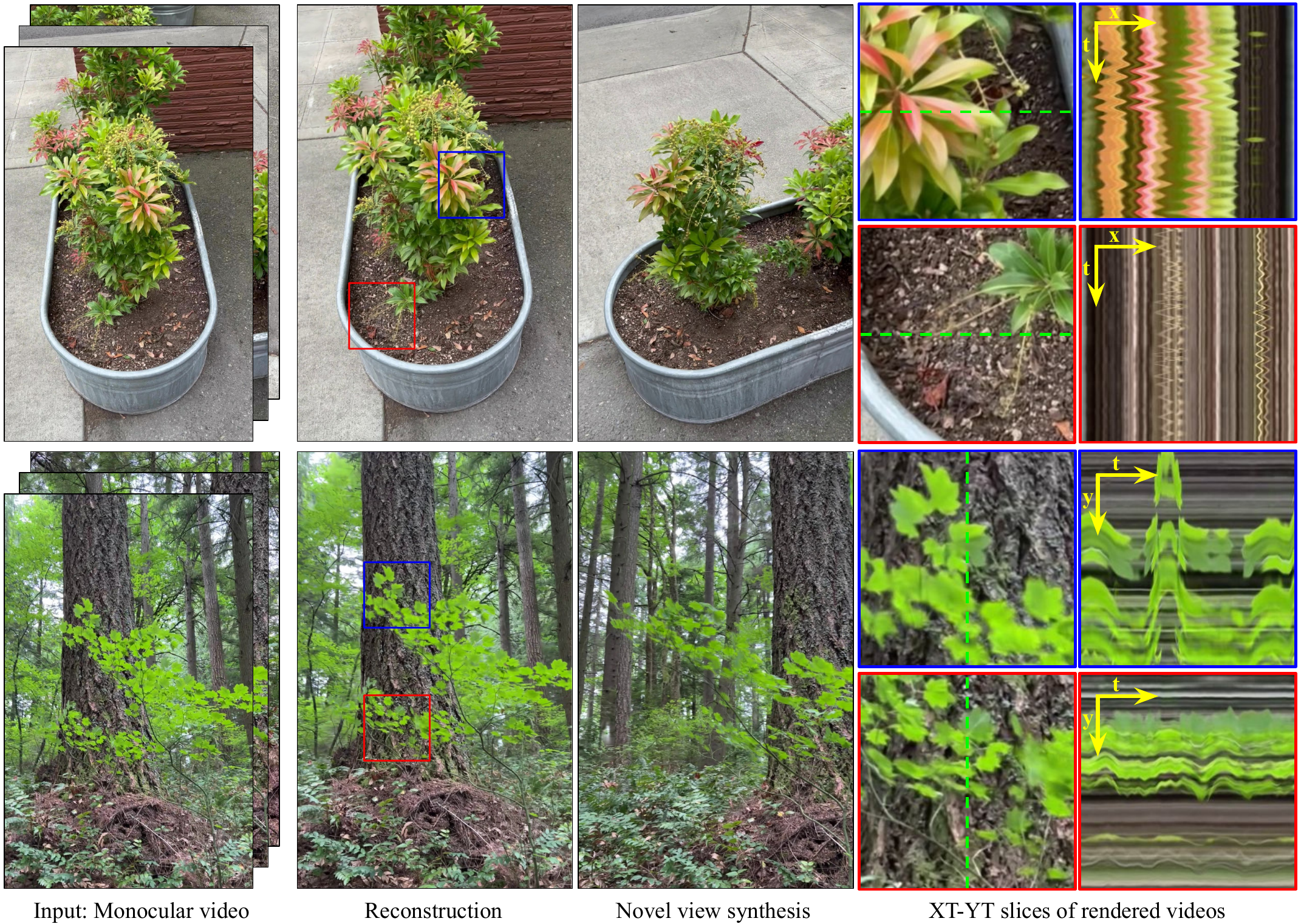}
  \caption{\textbf{Dynamic free-view synthesis of scenes with ambient motion.}
Given a causally captured video (1st col) in a scene with ambient dynamics, e.g., swaying trees in a forest or oscillating leaves of a plant in an outdoor environment, we reconstruct highly detailed 3D scene and the scene dynamics. 
Our method produces sharp reconstruction (2nd col) and render high-fidelity novel views at a specific time step (3rd col).
Our video rendering also capture complex ambient scene dynamics, providing immersive viewing experiences beyond navigating in purely static scenes (4th col).}
  \Description{}
  \label{fig:teaser}
\end{teaserfigure}

\maketitle

\section{Introduction}
\label{sec:intro}
We live in a dynamic world with rich and detailed textures and motions, where elements like leaves swaying rhythmically with the subtle influences of the wind and blades of grass are in constant, gentle motion.
The \emph{ambiance} makes the viewing experience immersive. Our work aims to achieve high-quality dynamic free-view synthesis from a \emph{monocular} capture of an ambient (plant) scene, providing a life-like viewing experience.
This has significant implications in areas such as
virtual 3D teleportation~\cite{orts2016holoportation} and virtual touring~\cite{broxton2020immersive}.

Dynamic view synthesis has seen considerable exploration through NeRF-based methods. These vary from learning static templates in canonical spaces~\cite{park2021nerfies, park2021hypernerf} to modeling motion fields across different timestamps~\cite{li2021neural,gao2021dynamic,liu2023robust}. 
However, these methods are limited by their training speed and overall reconstruction quality.
Recently, 3D Gaussian Splatting (3D-GS)\cite{kerbl3Dgaussians} has emerged as a powerful tool for reconstructing complex static scenes with high fidelity and fast training speeds. It represents the scene with 3D Gaussians. 
To render an image, we project the 3D Gaussians onto a 2D plane (splatting). 
The positions, rotations, sizes, colors, and opacities of these 3D Gaussians can be optimized by minimizing the photometric loss between the rendered image and the corresponding input image.
The extension of Gaussian Splatting to 4D has opened new possibilities for dynamic reconstruction, either through modeling the scene with 4D Gaussian primitives\cite{yang2023gs4d} or a set of 3D Gaussians~\cite{luiten2023dynamic}.
%
%
However, most of these methods depend on multi-camera captures~\cite{luiten2023dynamic}, a requirement impractical for most users.
A few methods can produce dynamic view synthesis from a monocular capture but are limited to bounded scenes~\cite{kratimenos2023dynmf,wu20234dgaussians}.
Additionally, these methods assume that all motion is observed; thus, they are unable to generalize to unseen motions.

\begin{figure}[t]
\newlength\figwidthLimitation
\setlength\figwidthLimitation{0.245\linewidth}
\centering%
\includegraphics[width=1\linewidth]{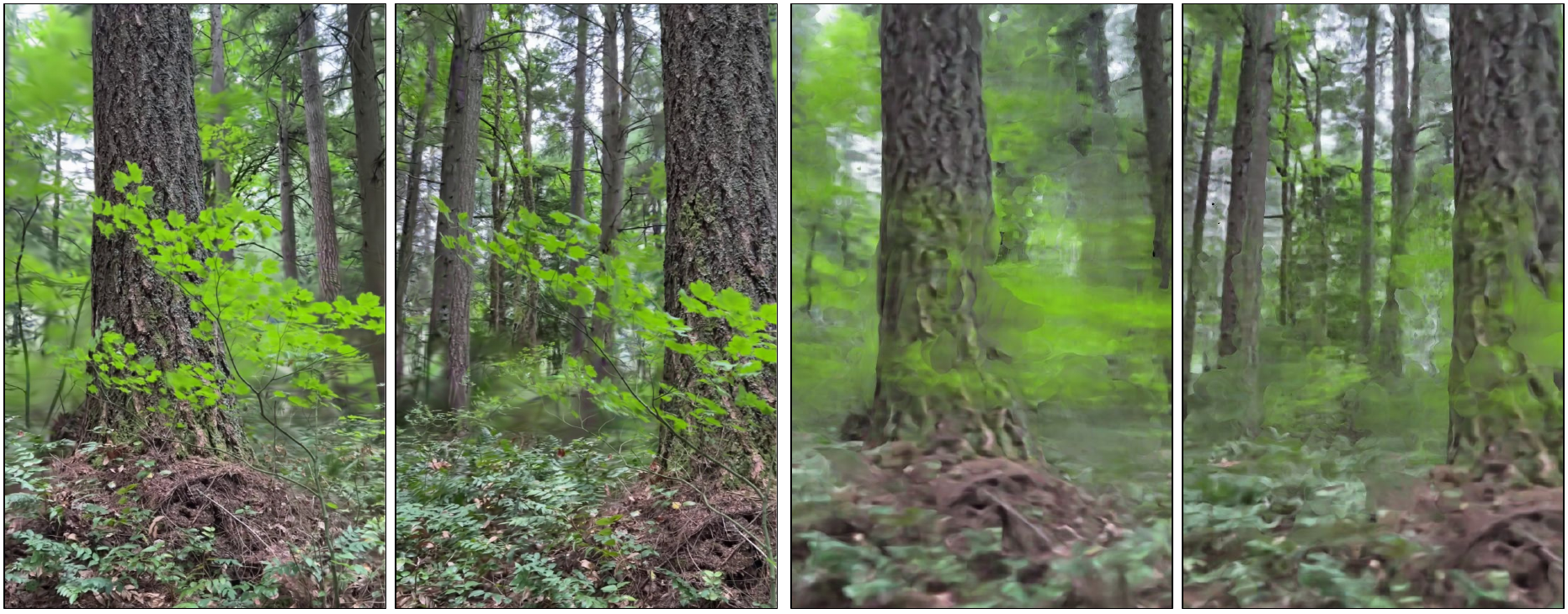}
\mpage{0.235}{\footnotesize{Reconstruction}}\hfill
\mpage{0.235}{\footnotesize{Novel view}}\hfill
\mpage{0.235}{\footnotesize{Reconstruction}}\hfill
\mpage{0.235}{\footnotesize{Novel view}}\hfill
\mpage{0.48}{$\underbrace{\hspace{\textwidth}}_{\substack{\vspace{-5.0mm}\\\colorbox{white}{~~4D-GS~\cite{wu20234dgaussians}~~}}}$}\hfill
\mpage{0.48}{$\underbrace{\hspace{\textwidth}}_{\substack{\vspace{-5.0mm}\\\colorbox{white}{~~RoDynRF~\cite{liu2023robust}~~}}}$}
\vspace{-3mm}
\captionof{figure}{
\textbf{Limitations of existing methods.} 
Here we highlight the limitations of state-of-the-art dynamic radiance fields in addressing the ambient dynamics in an unbounded scene.
RoDynRF suffers from severe blurriness due to the use of voxel-grid representation.
4D-GS can recover some spatial details for contents close to the cameras, but struggles with handling ambient motion, resulting in unstable foreground motion and inaccuracies in background motion. 
Please refer to the supplementary video for comparison.
}
\Description{}
\vspace{-0mm}
\label{fig:motivation}
\end{figure}

%
In this paper, we introduce a new algorithm for dynamic free-view synthesis from a monocular capture of a plant scene that addresses these limitations.
More specifically, we learn a set of template 3D Gaussians in canonical space and then learn how to deform these 3D Gaussians to model the time-varying volume density and appearance of the plant scene.
We fix the sizes, colors and opacities of the 3D Gaussians while learning time-dependent positions and rotations.
To model the motion, we propose to fit the observed motion of each 3D Gaussian to a \emph{trajectory}.
We find that various forms of periodic motion, especially those observed in nature, such as the motion of plants, can be effectively modeled using combinations of periodic functions and their associated coefficients.
Inspired by DCT-NeRF~\cite{wang2021neural}, we utilize basis functions derived from the discrete cosine transform (DCT) to represent the time-variant deformation. 
For each 3D Gaussian, we use a small MLP to predict the coefficients for each basis, with the final trajectory being the linear combination of all these bases.
This method enables us to \emph{extrapolate} motion trajectories beyond the observed data, thereby broadening the scope of our synthesis capabilities.

However, one of the primary challenges in this process is the network's ability to learn plausible deformations from just a single 2D image observation at each time step.
While infinitely many solutions can correctly render the input video, not all of them will result in plausible, photorealistic dynamic novel view synthesis.
We introduce rigidity regularization~\cite{luiten2023dynamic} to assist in learning a more consistent deformation in both spatial and temporal dimensions.

We validate our method's dynamic free-view synthesis performance on our proposed real-world plant scene dataset, demonstrating its effectiveness in synthesizing unseen motion.
Additionally, we showcase the capability of our method to edit motion, highlighting its versatility and practical applications.

Our \emph{contributions} are summarized as follows:
\begin{itemize}[noitemsep,topsep=0pt]
\item We introduce a high-quality dynamic free-view synthesis method from a monocular capture of unbounded ambient scenes.
\item We leverage motion trajectory to generalize to unseen motions.
\item We provide a real-world plant scene dataset featuring ambient motion. We demonstrate our method's capabilities in both motion editing and motion synthesis.
\end{itemize}
\section{Related Work}
\label{sec:related}
\begin{figure*}[t]    
\centering
\includegraphics[width=\linewidth]{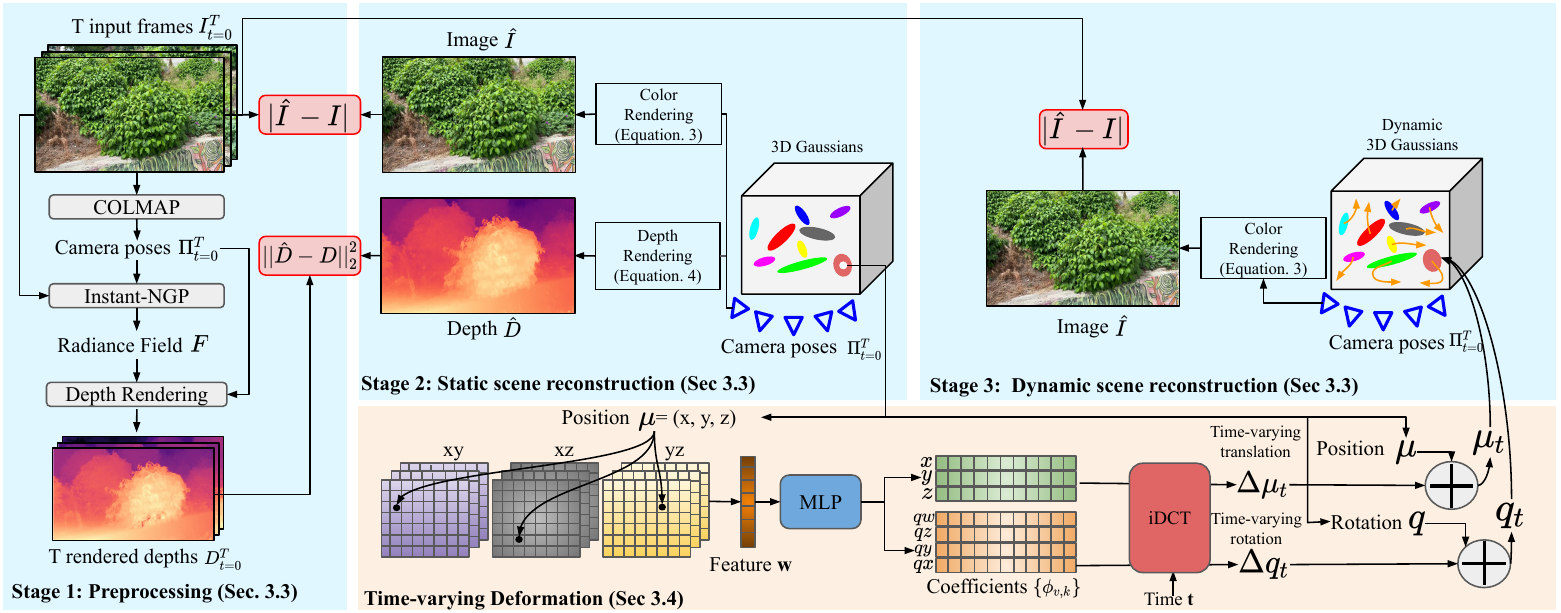}
\caption{\textbf{Method Overview.}
Our method comprises three stages: 1) pre-processing, 2) static scene reconstruction, and 3) dynamic scene reconstruction.
In pre-processing stage, we extract the rendered depth map using Instant-NGP~\cite{mueller2022instant}. 
The rendered depth from radiance field provide essential depth regualrization for unbounded scenes due to poor 3D point cloud recovery in distant regions. 
In the static scene reconstruction phase, we leverage both photometric and depth loss information obtained from the captured photos and reconstructed radiance field (from Instant-NGP), respectively. 
This stage allows us to produce high-quality static 3D Gaussian representations of the scene. 
However, the resulting representations do not model time-varying components like the ambient scene motion. 
The dynamic regions inevitably are blurry (due to inconsistent photometric losses across different frames).
In the dynamic scene reconstruction stage, we introduce temporal parameters to explicitly model the dynamics of each individual 3D Gaussians. 
We do so by using a triplane-based representation to predict the DCT coefficents for each point (the center position of each 3D Gaussian).
Using the predicted coefficients, we can recover the time-varying \emph{translation} and \emph{rotation} given any time step $t$. 
We supervise the motion representation using photometric loss and rigidity regularization. 
The resulting representation allow us to model the time-varying 3D Gaussians and thus render high-quality frames from novel view and time.
}
\Description{}
\label{fig:method-overview}
\end{figure*}

\paragraph{Novel view synthesis}
Novel view synthesis has been studied since decades ago. Earlier research relies~\cite{buehler2001unstructured, cayon2015bayesian, debevec2023modeling, gortler2023lumigraph, by2013plenoptic, kopf2014first, levoy2023light} on proxy geometry such as depth map to blend multiple source view and project to the target view. 
Beyond that, ~\cite{hedman2018deep} encodes deep features to the proxy geometry and decodes it with a learning-based encoder to provide better rendering quality. 
Besides, ~\cite{flynn2019deepview, srinivasan2019pushing, zhou2018stereo} exploit multi-plane images with different learning strategies, resulting in efficient and high-quality view synthesis in forward-facing scenarios.
A few years ago, NeRF~\cite{mildenhall2020nerf} proposed a way based on implicit field representation and used differentiable volumetric rendering to render images and update the implicit field. Since then, numerous works have been proposed to improve the view-synthesis technique in every aspect. 
For example, ~\cite{barron2021mip,barron2022mipnerf360,xu2022point,barron2023zip} are proposed to improve the quality in bounded and unbounded scenes and ~\cite{mueller2022instant,yu2021plenoctrees,fridovich2022plenoxels,sun2022direct} are proposed to improve training and inference speed. 
Recently, 3D Gaussian Splatting~\cite{kerbl20233d} introduces a novel approach to scene representation. 
This method leverages anisotropic 3D Gaussians as a representation for scenes and introduces an efficient differentiable rasterizer that exploit the splatting of these Gaussians onto the image plane. The key advantage of this technique is its ability to achieve fast, high-resolution rendering. However, these approach are limited to reconstruct and rendering static scene.
\paragraph{View synthesis of dynamic scenes.}
View synthesis of dynamic scenes is a challenging problem as it needs to deal with the reconstruction of geometry, appearance, and scene representation all at once. Earlier works~\cite{collet2015high,jain1995multiple,kanade1997virtualized,li2017robust,li20184d,yang2002real,BansalCVPR20,videoloop} tackle this problem by employing multiple-view camera setting to reconstruct the proxy geometry better and bake appearance on it. Beyond this, ~\cite{thonat2021video} uses a multiscale representation of motion that allows for looping and blending. Since the emergence of NeRF~\cite{mildenhall2020nerf}, numerous related papers~\cite{du2021neural,fridovich2023k,icsik2023humanrf,li2022streaming,li2022neural,lin2023im4d,shao2023tensor4d,song2023nerfplayer,wang2023mixed,wang2022fourier} make huge breakthrough on this domain. 
In addition, the requirement of dynamic view synthesis using only monocular video as input started to be fulfilled. For example,~\cite{li2021neural, gao2021dynamic, liu2023robust} rely on frame-to-frame correspondence to supervise the motion,~\cite{park2021nerfies, park2021hypernerf, TiNeuVox, Guo_2022_NDVG_ACCV,petitjean2023modalnerf} relies on the mapping from 3D space to a canonical space at each time-step to better fuse the information from multiple frames, and~\cite{wang2021neural, li2023dynibar} use trajectories to leverage the advantages of both sides.
In the past few months, there are a bunch of works~\cite{luiten2023dynamic,yang2023deformable3dgs,wu20234dgaussians,yang2023real,katsumata2023efficient,kratimenos2023dynmf,shao2023control4d,huang2023sc,das2023neural,lin2023gaussian,yu2023cogs,liang2023gaufre,li2023spacetime} start to exploit 3D Gaussian Splatting and applied it to the dynamic scene scenario. 
These works reach high-quality results in rendered images and fast inference time. However, challenges still exist, particularly when the input is monocular video instead of multiple view.
\paragraph{Modeling ambient motion}
People have used various techniques in the frequency domain to model natural, oscillatory 3D motion, such as the swaying of trees in the wind. 
For example,~\cite{davis2015image} shows how to use modal analysis in the frequency domain to describe the scene from input video and even simulate its motion.~\cite{petitjean2023modalnerf} shows that such technique can adapted to 3D scene.~\cite{li2023generative} shows that one can generate various plausible periodic motions by leveraging diffusion models.


\section{Method}
\label{sec:method}
We first introduce the background of Gaussian Splatting in Section~\ref{sec:preliminaries-3dgs}.
Following that, we provide an outline of our method in Section~\ref{sec:overview} with details.
Subsequently, we delve into the specifics of modeling periodic motion in Section~\ref{sec:dct-transform}.
Finally, in Section~\ref{sec:details}, we discuss strategies for mitigating memory consumption during training with many Gaussians along with other pertinent implementation details.

\subsection{3D Gaussian Splatting}
\label{sec:preliminaries-3dgs}
3D Gaussian Splatting (3D-GS)\cite{kerbl3Dgaussians} is a recently proposed explicit representation for scene reconstruction.
A set of 3D Gaussians represents the scene.
Each Gaussian $G$ is defined by a center point $\mu \in \mathbb{R}^3$ and a covariance matrix $\Sigma \in \mathbb{R}^{3x3}$,
composed of a rotation matrix $R \in \mathbb{R}^{3x3}$ and a 3-D scaling diagonal matrix $S \in \mathbb{R}^{3x3}$:
\begin{equation}
\label{eq:gaussian-dist}
    G(x|\mu;\Sigma) = e^{-\frac{1}{2}(x-\mu)^{T}\Sigma^{-1}(x-\mu)}
\end{equation}
\begin{equation}
\label{eq:covariance}
    \Sigma = RSS^TR^T
\end{equation}
Usually, the rotation matrix $R$ and scaling matrix $S$ are represented by the rotation unit quaternion $q$ and scaling factors $s \in \mathbb{R}^3$, respectively, for easier optimization.
In addition to Eq. (\ref{eq:gaussian-dist}) and (\ref{eq:covariance}) which define the basic properties of a 3D Gaussian, additional attributes are incorporated to enable photo-realistic rendering. These include color $c \in \mathbb{R}^3$ defined by coefficients of spherical harmonics (SH), and opacity $\alpha \in \mathbb{R}$.
To render an image, 3D-GS~\cite{kerbl3Dgaussians} first arrange these Gaussians based on their proximity to a designated viewpoint, then exploit the over-composite blending function to aggregate them for rendering:
\begin{equation}
\label{eq:blending-color}
    C = \underset{i \in N}{\sum}c_i \alpha_i \underset{j=1}{\overset{i-1}{\Pi}}(1-\alpha_i)
\end{equation}
The positions, rotations, sizes, colors, and opacities of these 3D Gaussians can be optimized by minimizing the photometric loss(i.e. L1 loss, SSIM loss) between the rendered image and the corresponding input image.


\subsection{Problem Statement}
We study the problem of dynamic view synthesis from monocular captures of ambient scenes.
Specifically, we focus on plant scenes, where the magnitude of motion is typically confined to a specific range, such as the swaying motion of trees.
While the design of 3D Gaussian splatting handles static scenes effectively, further adaptations are necessary to accommodate the temporal dynamics of scenes. Specifically, properties such as the center position $\mu$ and rotation $q$ must be made time-dependent and are denoted as $\mu_t$, $q_t$, respectively (see ~\ref{sec:overview} for detail). Here, the subscript $t$ is the time index.

\begin{figure}[t]
\newlength\figwidthDepth
\setlength\figwidthDepth{0.33\linewidth}
\centering%
\parbox[t]{1\linewidth}{\centering%
 \includegraphics[width=0.9\linewidth]{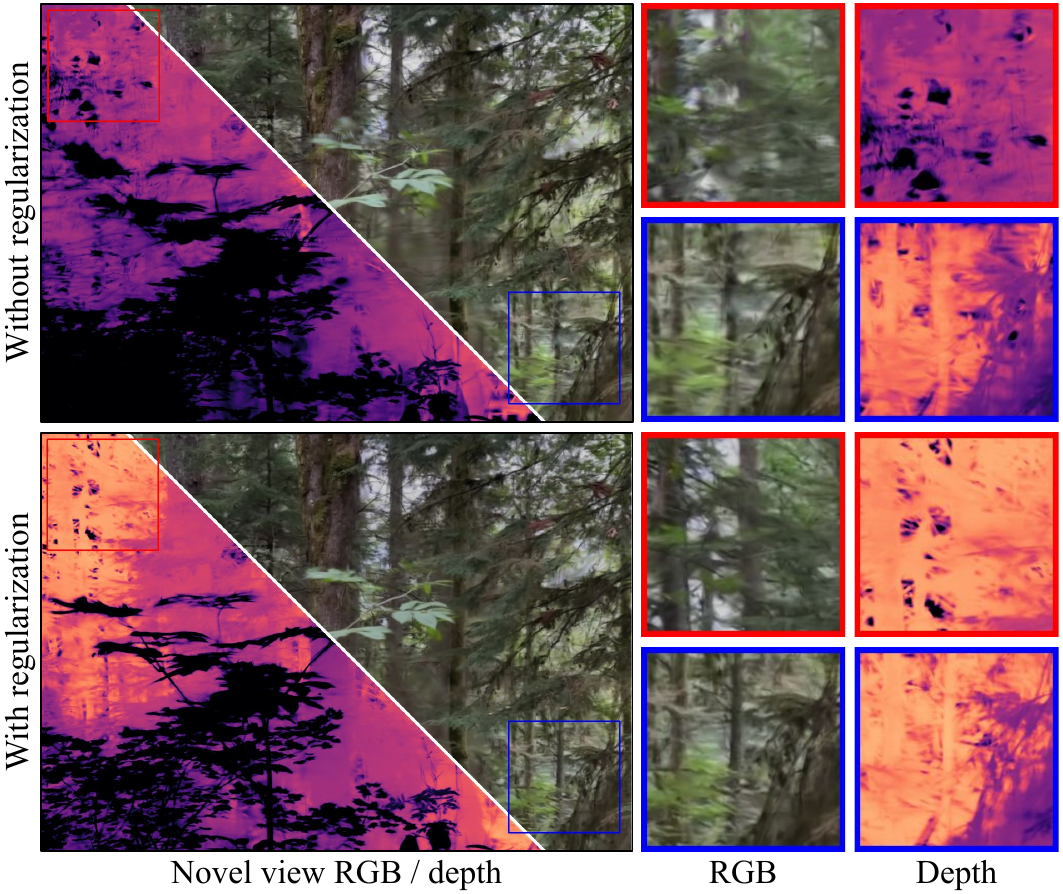}}
\vspace{-6mm}
\captionof{figure}{
\textbf{The importance of depth regularization.} 
The quality of 3D-GS depends heavily on the accurate 3D point cloud initialization.
In unbounded scenes, however, the geometry of scene elements far away from the camera cannot be reliably reconstructed with structure from motion algorithms (due to small motion parallax). 
Consequently, 3D-GS tends to predict incorrect geometry in the background and render blurry image due to the lack of initial Gaussians (top). 
We address this challenge by applying the depth regularization. 
With the regularization, we observe more accurate and detailed appearance in our rendering (bottom).
}
\Description{}
\label{fig:nerf_depth}
\vspace{-0mm}
\end{figure}

\subsection{Method Overview}
\label{sec:overview}
We show our proposed framework in Fig.~\ref{fig:method-overview}. 
Given an input video sequence of $T$ frames and the corresponding camera parameters, we designed a three-stage approach to reconstruct the dynamic ambient scene using a collection of temporal 3D Gaussians. These Gaussians serve as comprehensive representations, capturing the scene’s geometry, appearance, and dynamic motion.
The three stages of our approach are \textbf{pre-processing}, \textbf{static scene reconstruction}, and \textbf{dynamic scene reconstruction}.

\paragraph{Pre-processing}
We first run COLMAP on the video frames $I_{i=0}^{T}$ to obtain the corresponding camera poses $\Pi_{i=0}^{T}$ and a sparse point cloud. These points help initialize a set of 3D Gaussians.
We have observed a strong correlation between the quality of 3D Gaussian splatting and the adequacy of the initial point coverage.
This connection stems from the nature of 3D Gaussian splatting, which relies on an explicit representation. 
If COLMAP fails to extract feature points in certain regions, it results in a lack of Gaussians to begin with.
To mitigate this issue, 3D-GS~\cite{kerbl3Dgaussians} introduced split and clone strategies, gradually enabling Gaussians to expand into poorly covered areas. While this approach shows promise in reconstructing content close to the initial points, it faces challenges in reconstructing content farther from them. 
The absence of Gaussians in areas with insufficient initial point coverage leads to noticeable artifacts and incorrect geometry (Fig.~\ref{fig:nerf_depth}), especially in the large unbounded scenes that are the focus of this paper.

Thus, we propose to leverage an implicit representation, specifically NeRF~\cite{mildenhall2020nerf}, to constrain the geometry further. 
Unlike 3D Gaussian splatting, NeRF and its subsequent developments~\cite{barron2022mipnerf360, barron2023zip, barron2021mip, mueller2022instant} utilize an implicit representation, allowing for arbitrary sampling of points across the 3D space.
As a result, it achieves plausible depth estimation in the under-covered region in the unbounded scene.
Leveraging the capabilities of NeRF, we fit an NGP~\cite{mueller2022instant} to the unbounded scene in just 10 minutes and subsequently render depth maps $D$ for each input frame.

One way to use the depth $D$ is to sample additional 3D points through depth unprojection and use them to initialize more 3D Gaussians. However, this can lead to excessive points for rendering, causing computational strain and out-of-memory issues. Therefore, we propose a more effective depth regularization.
In the following paragraph, we will detail how these depth maps guide the Gaussians, thus extending their influence into regions considerably distant from the initial points.


%

\paragraph{Static scene reconstruction}
We focus on plant scenes, where the magnitude of motion is typically confined to a specific range, such as the swaying motion of trees. Moreover, a significant portion of our scene remains static, e.g., rocks, land, and tree trunks.
Therefore, we begin by assuming the absence of motion within the scene, intending to introduce motion later in the process.
This approach offers the advantage of breaking down the complex task of simultaneously reconstructing geometry, appearance, and motion into two more manageable sub-problems: first reconstructing geometry and appearance in the canonical space, and subsequently addressing the reconstruction of motion.
As mentioned previously, 3D-GS heavily depends on COLMAP 3D point cloud initialization. The absence of 3D points in under-covered regions will lead to incorrect geometries.
To solve this issue, we introduce a depth regularization. 
By replacing color $c_i$ with depth value of Gaussian, $z_i$ in Eq.~\ref{eq:blending-color}, we can render depth map:
\begin{equation}
\label{eq:blending-depth}
    \hat{D} = \underset{i \in N}{\sum}z_i \alpha_i \underset{j=1}{\overset{i-1}{\Pi}}(1-\alpha_i)
\end{equation}
Similar to the photometric loss, we acquire depth regularization through the L2 loss between the depth map $\hat{D}$ generated by Gaussian splatting and NeRF-depth $D$ in the pre-processing stage.
This regularization effectively improves the reconstruction quality, as shown in Fig.~\ref{fig:nerf_depth}.
The large unbounded scenes often significantly increase the number of Gaussians, making training expensive and less efficient.
To address this issue, we employ a mask loss and mask pruning technique~\cite{lee2023compact} to reduce the number of Gaussians used in our model effectively.
Furthermore, we implement gradient accumulation and develop a multi-pass rendering scheme for memory-efficient training, which we detail in Sec.~\ref{sec:details}.

\begin{figure}[t]
\newlength\figwidthContraction
\setlength\figwidthContraction{0.331\linewidth}
\centering%
\parbox[t]{1\linewidth}{\centering%
 \includegraphics[width=1.0\linewidth]{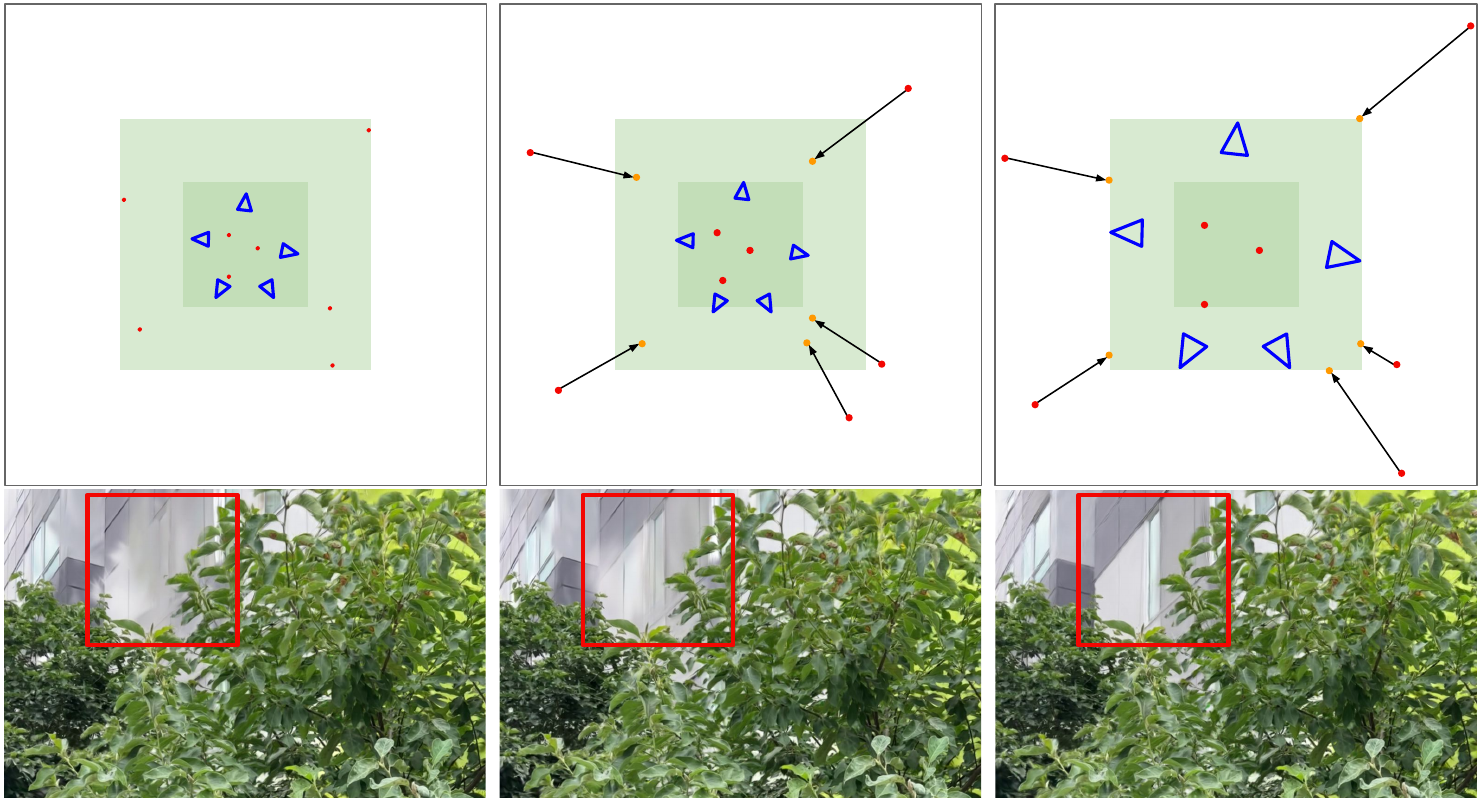}}
\mpage{0.32}{\footnotesize{(a) Normalize by point cloud}}\hfill
\mpage{0.32}{\footnotesize{(b) Normalize by camera poses then contract}}\hfill
\mpage{0.32}{\footnotesize{(c) Normalize by camera poses then map}}\hfill
\vspace{-3mm}
\captionof{figure}{
\textbf{Effect of different  scene normalization strategies.}
(a) Normalizing the scene based on the \emph{range of 3D point cloud} often results in an inefficient use of representational power (because the scene scale can be very large). 
This typically leads to a blurry foreground and an incorrectly rendered background.
(b) When normalizing the scene using the \emph{range of camera poses} and applying $\infty$-norm contraction, the foreground becomes sharper. However, the background remains blurry due to inaccurately predicted motion.
(c) We propose to normalize the scene using the range of camera poses and map points outside this range to the boundary. Our results show that this achieves higher-quality synthesis in both the foreground and background regions.
}
\Description{}
\vspace{-0mm}
\label{fig:scene_contraction}
\end{figure}

\paragraph{Dynamic scene reconstruction}
After achieving a satisfying static reconstruction in the canonical space, we now focus on bringing back the motion to model the dynamic scene. For each 3D Gaussian, we want to predict a time-dependent translation $\Delta \mu_{t}$ and 
rotation quaternion $\Delta q_{t}$ to deform it from the canonical space to a time-dependent space at time $t$.
Specifically, the center position $\mu$ and rotation $q$ become $\mu_t$ and $q_t$, where:
\begin{equation}
\label{eq:time-variant-translation}
    \mu_{t} = \mu + \Delta \mu_{t}
\end{equation}
\begin{equation}
\label{eq:time-variant-rotation}
    q_t = \frac{(q + \Delta q_{t})}{||q + \Delta q_{t}||}
\end{equation}
However, instead of directly predicting $\Delta \mu_{t}$ and $\Delta q_{t}$, we propose to first model a motion \emph{trajectory}, and then sample time-varying $\Delta \mu_{t}$ and $\Delta q_{t}$ by querying the trajectory at time-step $t$.
This novel design enables us to extrapolate motion trajectories beyond the observed data, thereby synthesizing unseen motions.

To model the motion trajectory of a 3D Gaussian centered at $\mu$, we encode the position using a voxel-based encoder~\cite{Cao2023HexPlane}. 
Unlike the position $\mu$, which is unbounded, the voxel-based encoder has a predefined range, $[-1, 1]$.
We have to normalize the position $\mu$ to fit this range.
We have observed that normalizing the scene based on the range of point clouds can lead to inefficient use of representational power on the static background. This often results in a blurry foreground and an incorrectly moving background, as depicted in Fig.~\ref{fig:scene_contraction}(a). 
We experiment with the $\infty$-norm contraction proposed in Mip-NeRF-360~\cite{barron2022mipnerf360}, specifically designed for unbounded scenes. However, the background remains blurry due to inaccurately predicted motion.
Therefore, we propose normalizing the scene based on the range of the input camera positions, ensuring all camera positions fall within the $[-1, 1]$ range.
We reduce their resolution for Gaussians outside this range by mapping them to the boundary.
This strategy preserves higher resolution for those salient motions in the foreground because the range of camera positions is typically smaller than that of the initial point cloud. Additionally, it serves as a \emph{regularization} for motions further from the camera, where significant motion is less expected.
This approach effectively achieves high-quality representation in both foreground and background regions, as demonstrated in Fig~\ref{fig:scene_contraction}(c).
The encoder output is then passed through an MLP to predict coefficients for the pre-defined basis. We take the linear combination of all these bases to obtain the final trajectory. 
Please refer to Sec.~\ref{sec:dct-transform} and Fig.~\ref{fig:method-overview} for a detailed explanation of how we model the trajectory.

\begin{figure}[H]
\setlength\figwidthLimitation{0.49\linewidth}
\centering%
\hfill
\parbox[t]{\figwidthLimitation}{\centering%
  \includegraphics[trim=0 0 0 150, clip=true, width=\figwidthLimitation]{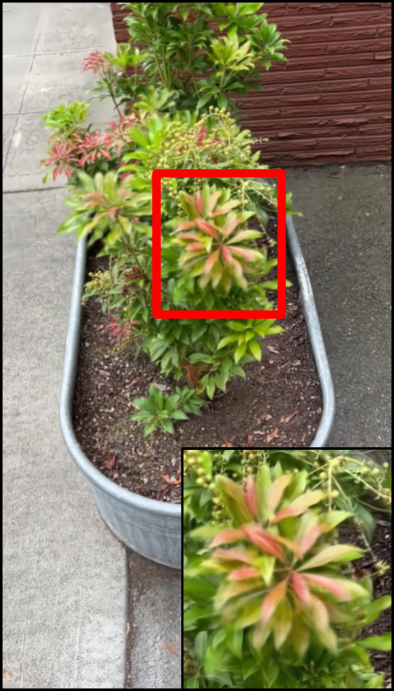}\\%
   Stage 2: Static scene reconstruction}%
\hfill%
\parbox[t]{\figwidthLimitation}{\centering%
  \includegraphics[trim=0 0 0 150, clip=true, width=\figwidthLimitation]{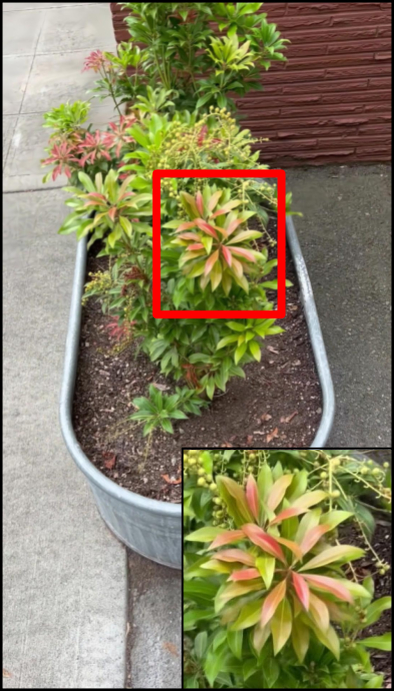}\\%
   Stage 3: Dynamic scene reconstruction}%
\hfill%
\captionof{figure}{
\textbf{Effect of time-varying parameters.} In stage 2, the absence of time-varying parameters leads to blurry and ghosting artifacts. In contrast, stage 3's joint optimization of time-varying and time-independent parameters allows for accurate reconstruction of ambient motion and 3-D geometry.
}
\Description{}
\label{fig:staticNdynamic}
\end{figure}

By jointly optimizing the time-varying deformation along with their static counterparts within the Gaussians, as outlined in Eq. \ref{eq:time-variant-translation} and \ref{eq:time-variant-rotation}, we can determine the status of the Gaussians at different time steps.
Then, we follow the same rasterization pipeline in 3D-GS~\cite{kerbl20233d} to render the image at time $t$.
The encoder and MLP are optimized by minimizing the photometric loss(i.e., L1 loss) between the rendered image and the corresponding input image.
After the training of dynamic scene reconstruction, the blurry and ghosting artifacts due to the inability to model ambient motion in static scene reconstruction is greatly reduced (see Fig.~\ref{fig:staticNdynamic}).

\subsection{Modeling periodic motion}
\label{sec:dct-transform}
Various forms of periodic motion, especially those observed in nature, such as the motion of plants, can be effectively modeled using periodic functions and their related coefficients.
Inspired by DCT-NeRF~\cite{wang2021neural}, we utilize basis functions derived from the discrete cosine transform (DCT) to represent the time-varying deformation (Fig.~\ref{fig:method-overview}).
We pre-define a set of DCT basis functions and predict the coefficients for each basis.
Each 3D Gaussian's time-varying deformation is independently modeled:
\begin{equation}
\label{eq:idct-transform}
    v(t) = \sqrt{\frac{2}{K+1}}\underset{k=1}{\overset{K}{\sum}}\phi_{v, k}\cos(\frac{\pi}{2T}(2t+1)k)
\end{equation}
Here, $\phi_{v, k}$ represents the $k$-th coefficient associated with the time-varying deformation $v$, while $v(t)$ represents any time-varying deformation scalar at time step $t$. For example, $v(t)$ could be the translation $\Delta \mu_{t}$ in the x-axis.
Utilizing DCT to represent time-varying deformation in our scenario offers two distinct advantages: storage efficiency and improved generalization capabilities. As highlighted in concurrent research~\cite{katsumata2023efficient}, representing time-varying deformation with basis functions enhances storage efficiency. In all of our experiments, we set $K = \ceil{\frac{1}{4}T}$, and significantly reduce the number of deformation parameters to only a quarter of the total.
Moreover, thanks to the inherent characteristics of DCT, this approach accommodates motion fitting when motion is discernible in the input images, and it replicates motion in the absence of direct supervision (e.g., when a Gaussian is occluded or out of the camera frustum). Given that the motion in our scenarios is also periodic (repetitive), the natural properties of DCT contribute to enhanced generalization capabilities for time-varying deformation.

We have observed that neighboring Gaussians often exhibit spatially inconsistent deformations, resulting in unnatural motion during dynamic novel view synthesis.
To tackle this issue, we leverage rigidity regularization~\cite{luiten2023dynamic}.
The movement of each 3D Gaussian $G$'s neighbors should adhere to the rigid-body transformation of $G$'s coordinate system across time steps.
Additionally, the neighbors of each 3D Gaussian $G$ should maintain similar rotations over time to ensure spatial and temporal coherence.



\subsection{Memory efficient multi-pass rendering}
While we have reduced the number of Gaussians in static scene reconstruction, GPU out-of-memory issues persist in unbounded scenes due to the optimization demands of the time-varying MLP. As we cache features at the MLP's hidden layers for each input sample (i.e. each Gaussian), we face significant memory usage. We employ a \emph{two-pass} rendering approach to handle it. In the first pass, we avoid intense memory usage by not caching any feature and proceed to obtain the time-varying deformation parameters from the MLP. We then deform the Gaussians according to these parameters, rasterize them, and identify those Gaussians involved in rendering the current image. We store the deformed status of the involved Gaussians, which will be used in the next pass.

In the second pass, we obtain the deformation using the MLP again. Two things are different here: First, we only consider the Gaussians identified in the previous pass to reduce memory usage. Second, we cache features for back-propagation. We then deform, rasterize, and render the involved Gaussians into an image. Finally, we compute the loss, back-propagate to obtain gradients, release the cache, and optimize the MLP.

To further address the memory constraints, in the second pass, we divide the involved Gaussians into chunks, each containing up to 500k samples. Sequentially, we cache features corresponding to each chunk for back-propagation and release them once we have obtained the gradients. This approach, known as gradient accumulation, circumvents the memory bottleneck. One issue remains: each time we only have a certain chunk of deformed Gaussian status and cannot render a complete image. Thus, we simply take the deformed status of the other chunks from the first pass. With these strategies, we resolve the memory issue.

\label{sec:details}
\section{Experimental Results}

\subsection{Experimental Setup}

\paragraph{Hyperparameters.}
We implement our approach using a single NVIDIA A100 GPU with 40GB of memory. 
During the pre-processing stage, we start by training the Instant-NGP model for 10k iterations.
In the static scene reconstruction stage, we train static 3D Gaussians for 80k iterations.
In the first 9k iterations, we use the split and clone operation to enhance the density of Gaussians. Subsequently, from the 9k to 80k iterations, we deactivate the densification operation and switch to the mask pruning method~\cite{lee2023compact} to reduce the number of Gaussians while preserving the visual quality of rendered images. The training time of this stage is about $1.5$ hours.
Moving on to dynamic scene reconstruction, we jointly train the MLP and 3D Gaussians through 30k iterations. The overall training procedure exhibits a variable duration, ranging from 1.5 to 8 hours, contingent on the complexity and scale of the scene. 

\subsection{Dataset}
Due to the scarcity of the unbounded real-world ambient scene dataset, we introduce the Forest dataset.
This dataset includes 10 dynamic ambient scenes captured in various environments. Each video lasts between 10 and 30 seconds, recorded at 1080p resolution and 30 fps. We use COLMAP~\cite{schonberger2016structure} to reconstruct the sparse point cloud and obtain camera parameters.
\subsection{Baseline}
We adopt two different baselines that have demonstrated their ability to handle unbounded scenes. The first one is Robust-DyNeRF (RoDynRF), which leverages the scene contraction technique from \cite{barron2022mipnerf360} to deal with unbounded dynamic scenes. The second one is 4D Gaussian Splatting (4D-GS)~\cite{wu20234dgaussians}, which also shows its ability to render high-quality images of dynamic scenes efficiently and outperforms opponents that use NeRF representation. We modify the default settings in 4D-GS by disabling the time-variant scaling parameters $\Delta s_t$, which we found to be unstable and prone to producing NaN issues.

\subsection{Qualitative Comparison}
\label{sec:visual_comparison}
\begin{figure*}
\newlength\figwidthNovel
\setlength\figwidthNovel{0.165\linewidth}
\centering%
\parbox[t]{\figwidthNovel}{\centering%
  \includegraphics[trim=0 0 0 0, clip=true, width=\figwidthNovel]{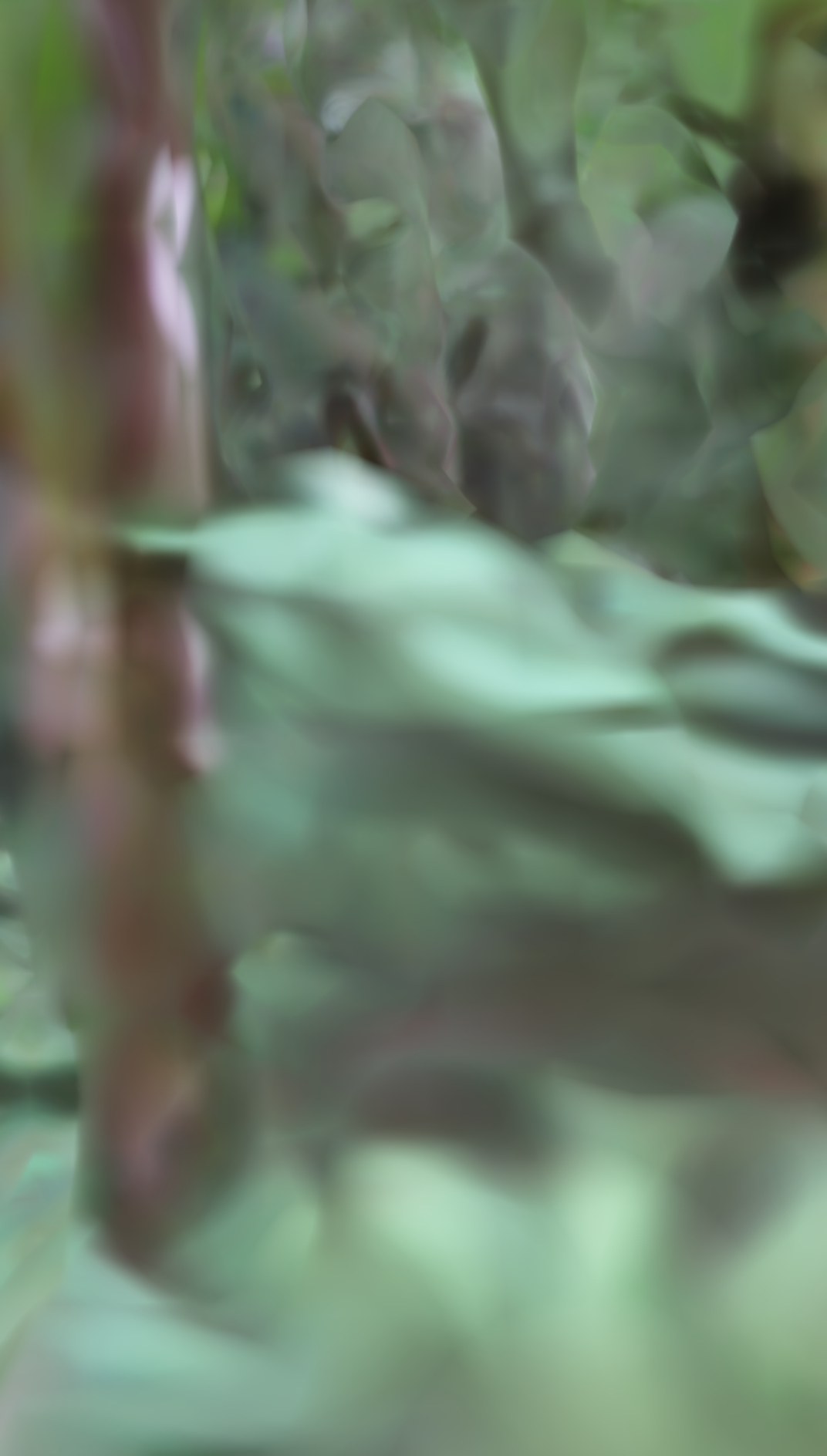}\\%
  \includegraphics[trim=0 0 0 0, clip=true, width=\figwidthNovel]{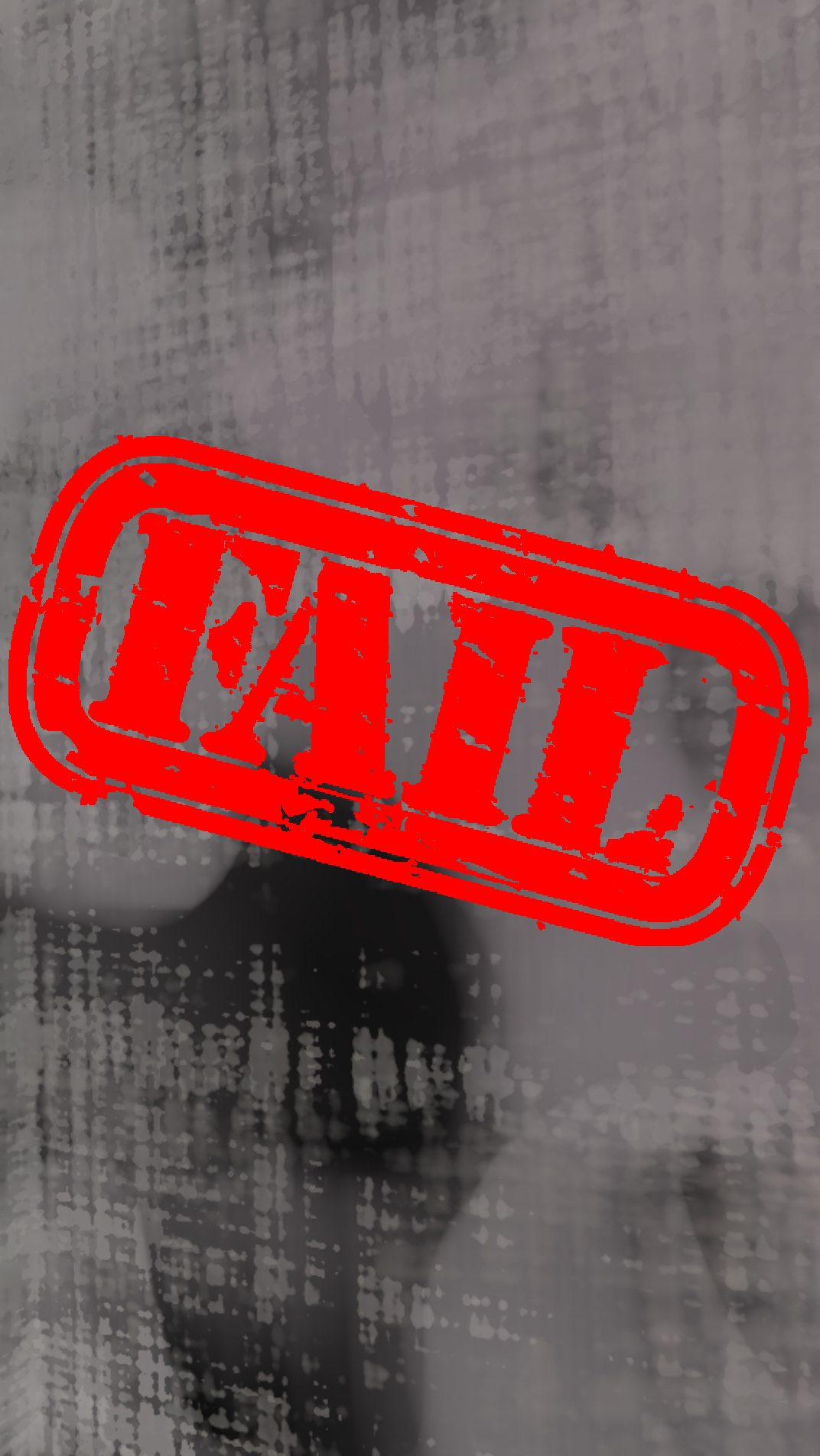}\\%
   RoDynRF}%
\hfill%
\parbox[t]{\figwidthNovel}{\centering%
  \includegraphics[trim=0 0 0 0, clip=true, width=\figwidthNovel]{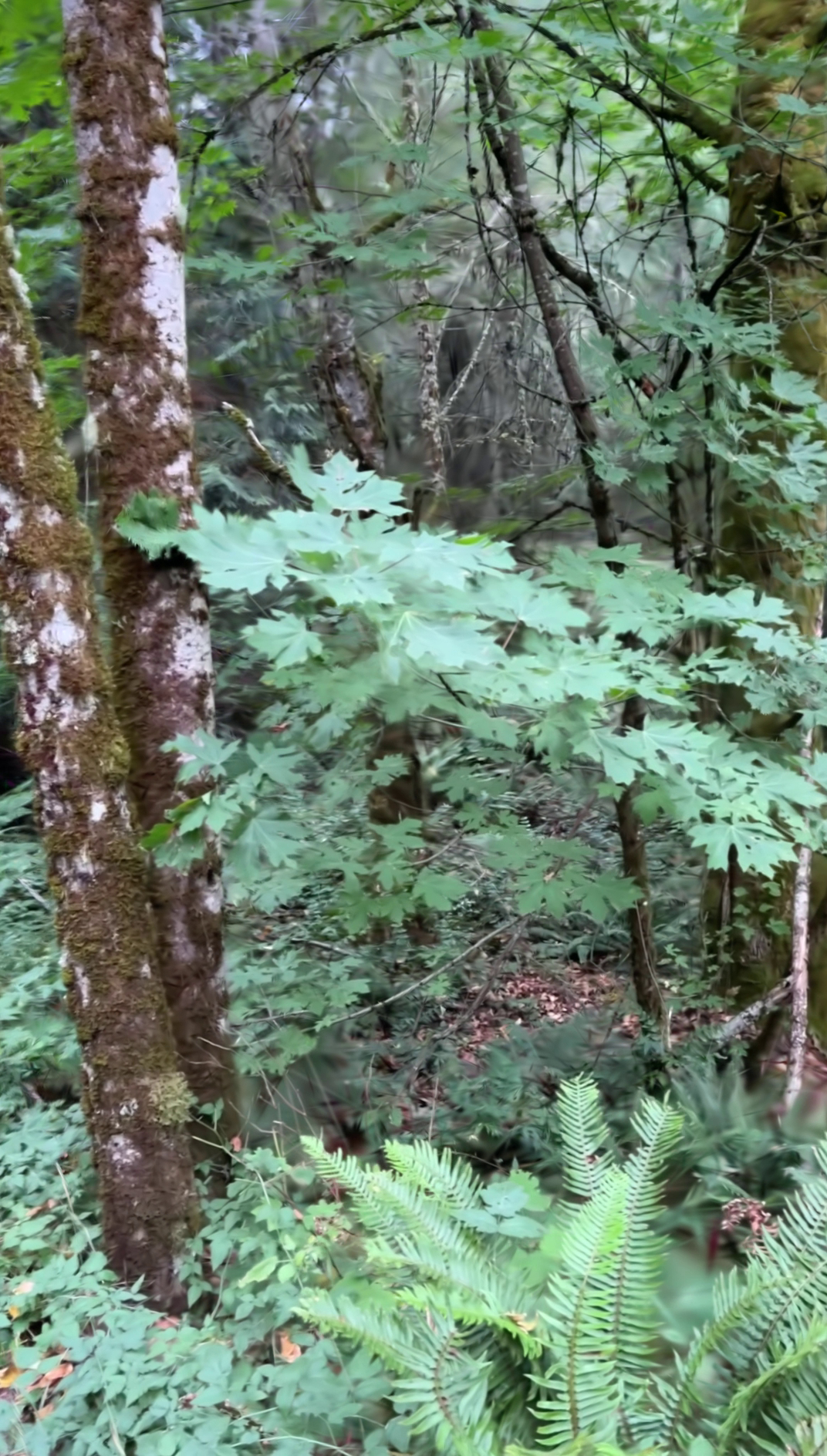}\\%
  \includegraphics[trim=0 0 0 0, clip=true, width=\figwidthNovel]{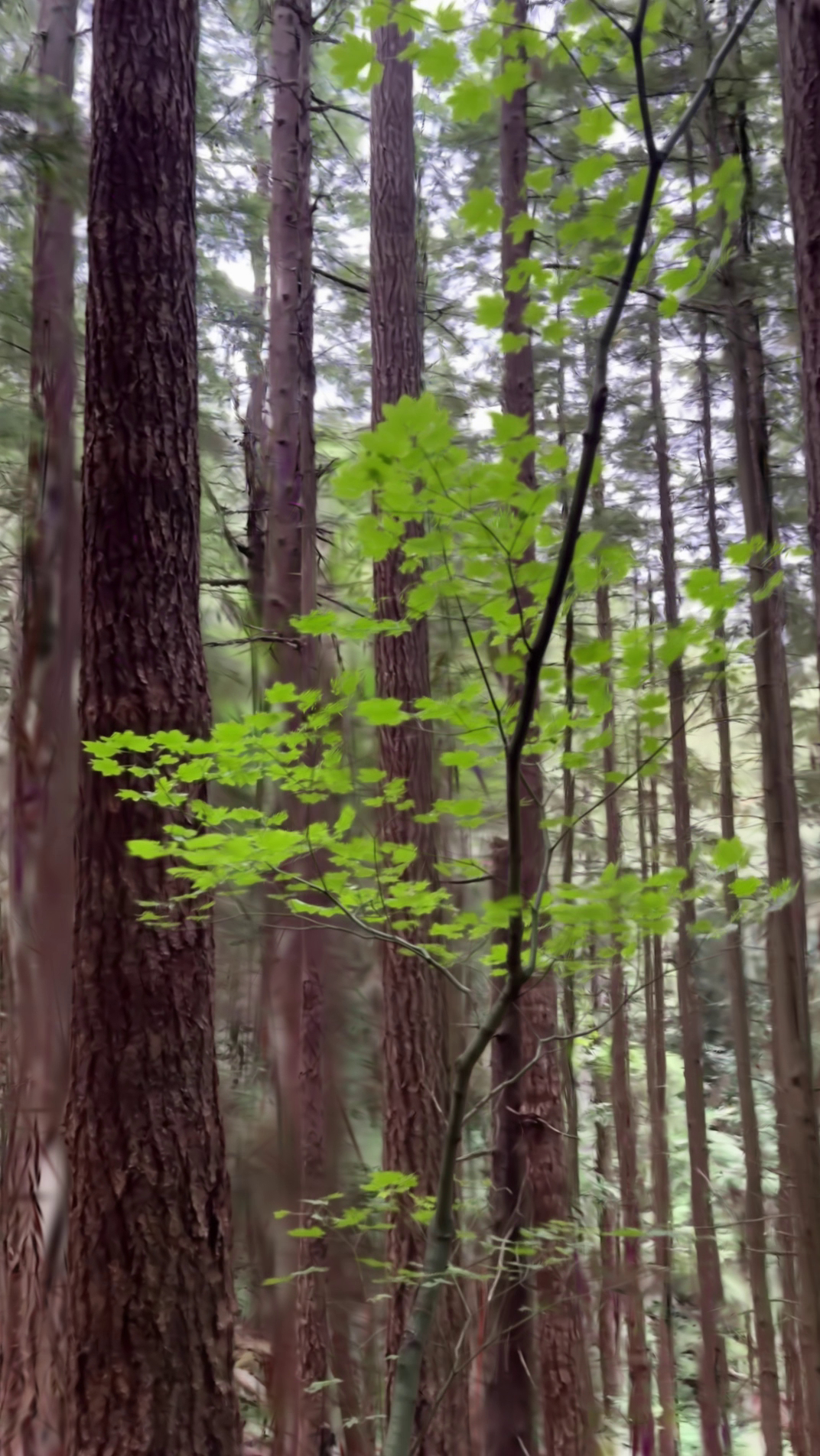}\\%
   4D-GS}%
\hfill%
\parbox[t]{\figwidthNovel}{\centering%
  \includegraphics[trim=0 0 0 0, clip=true, width=\figwidthNovel]{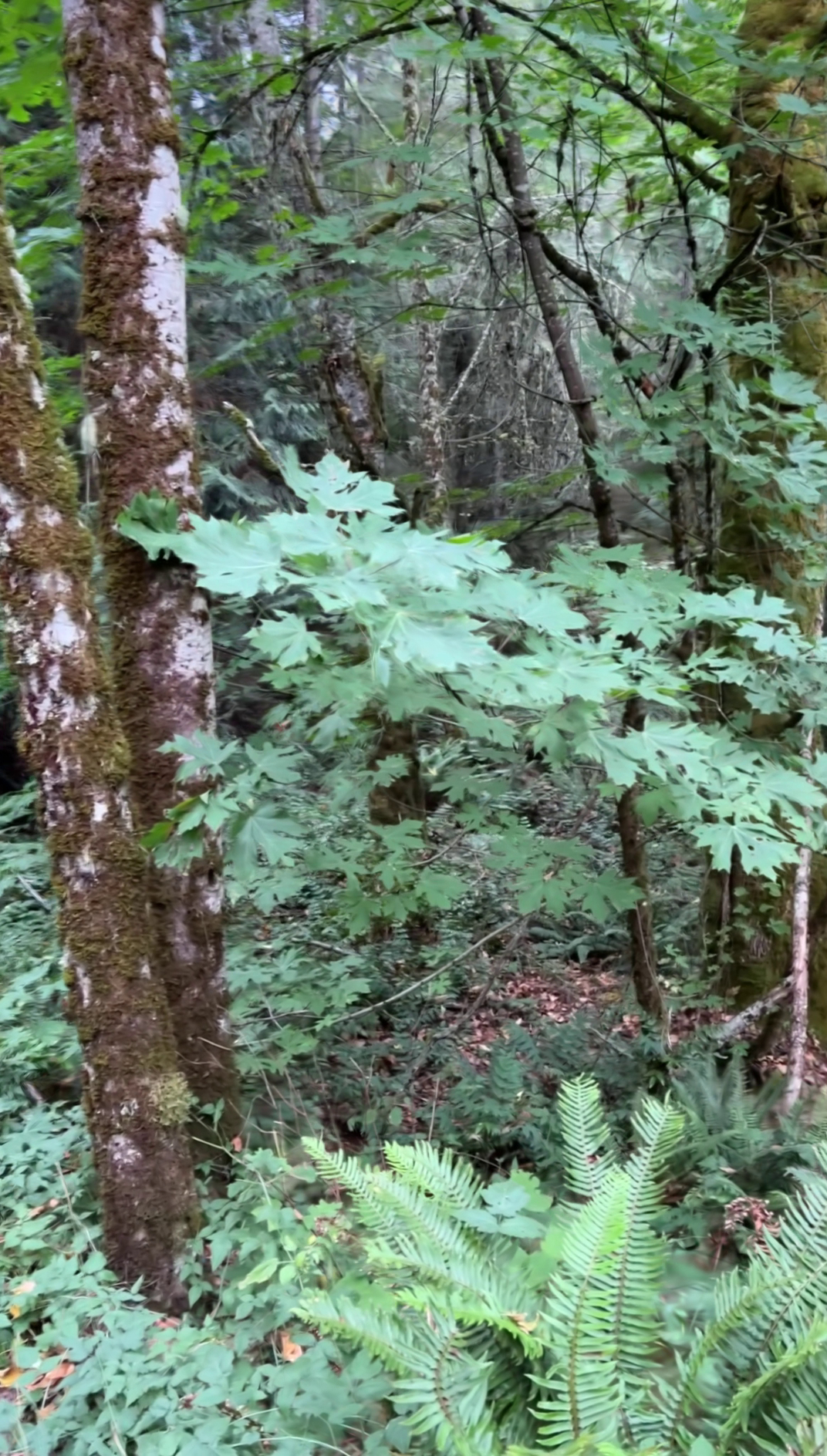}\\%
  \includegraphics[trim=0 0 0 0, clip=true, width=\figwidthNovel]{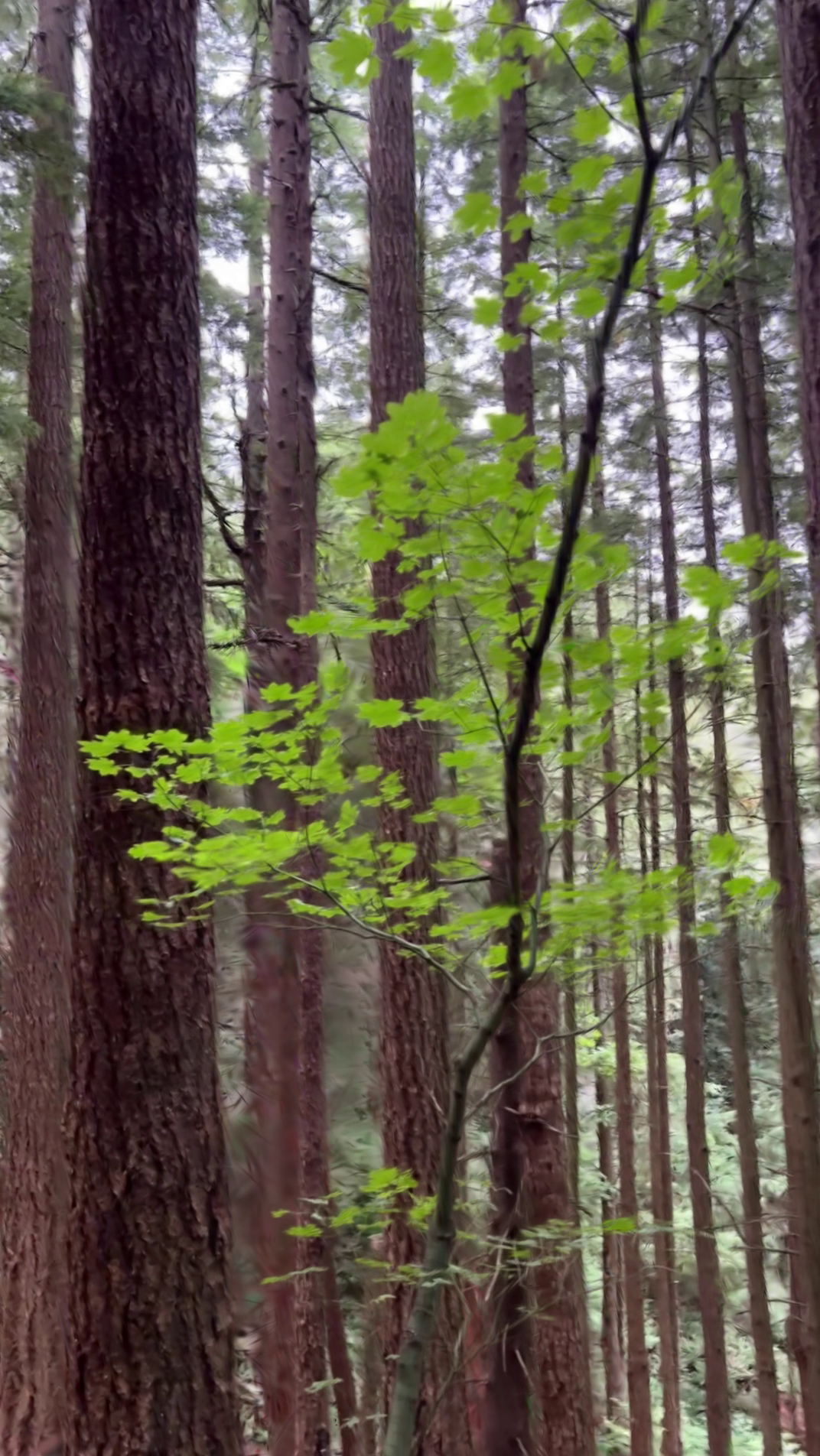}\\%
   Ours}%
\hfill%
\parbox[t]{\figwidthNovel}{\centering%
  \includegraphics[trim=0 0 0 0, clip=true, width=\figwidthNovel]{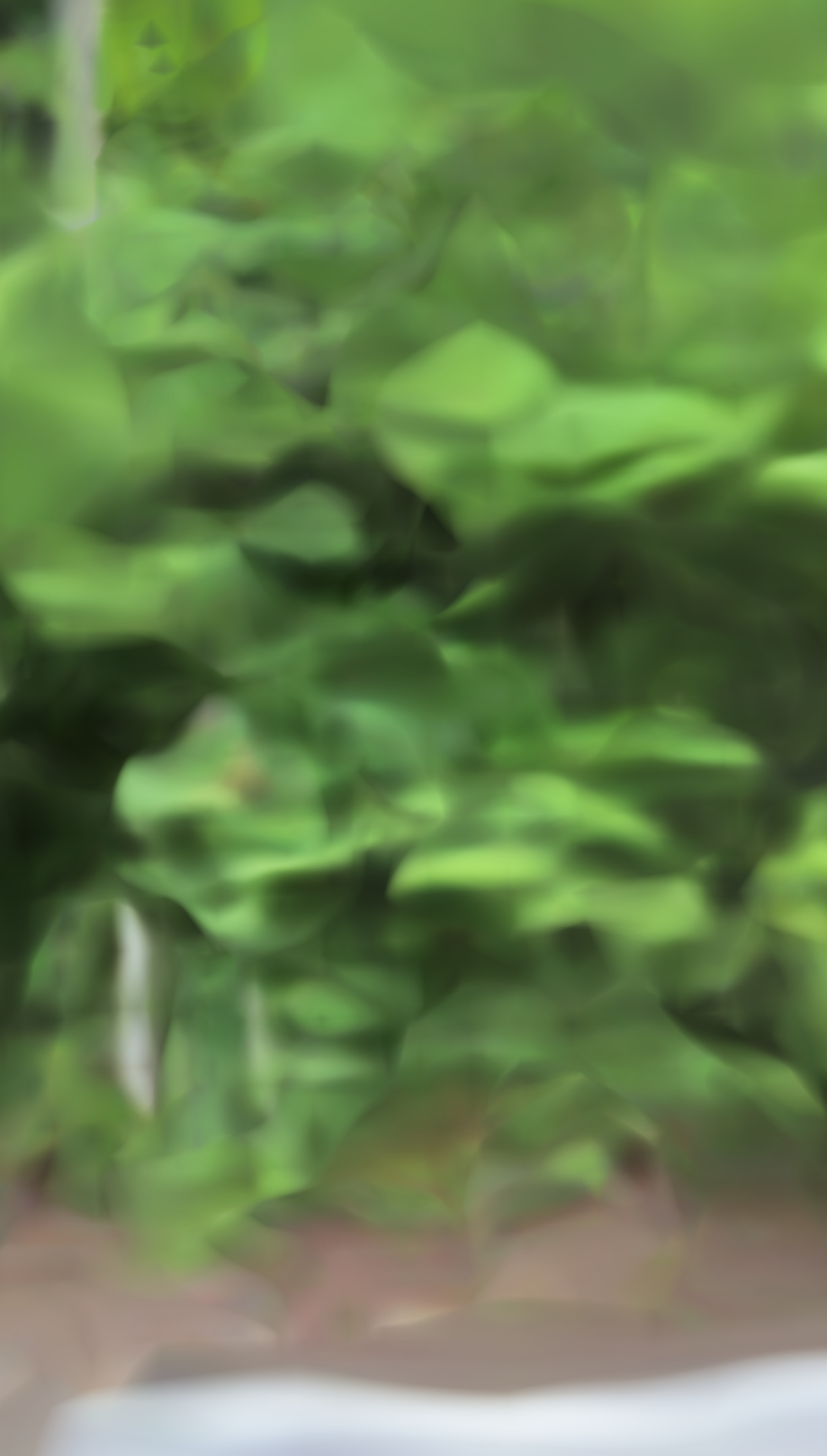}\\%
  \includegraphics[trim=0 0 0 0, clip=true, width=\figwidthNovel]{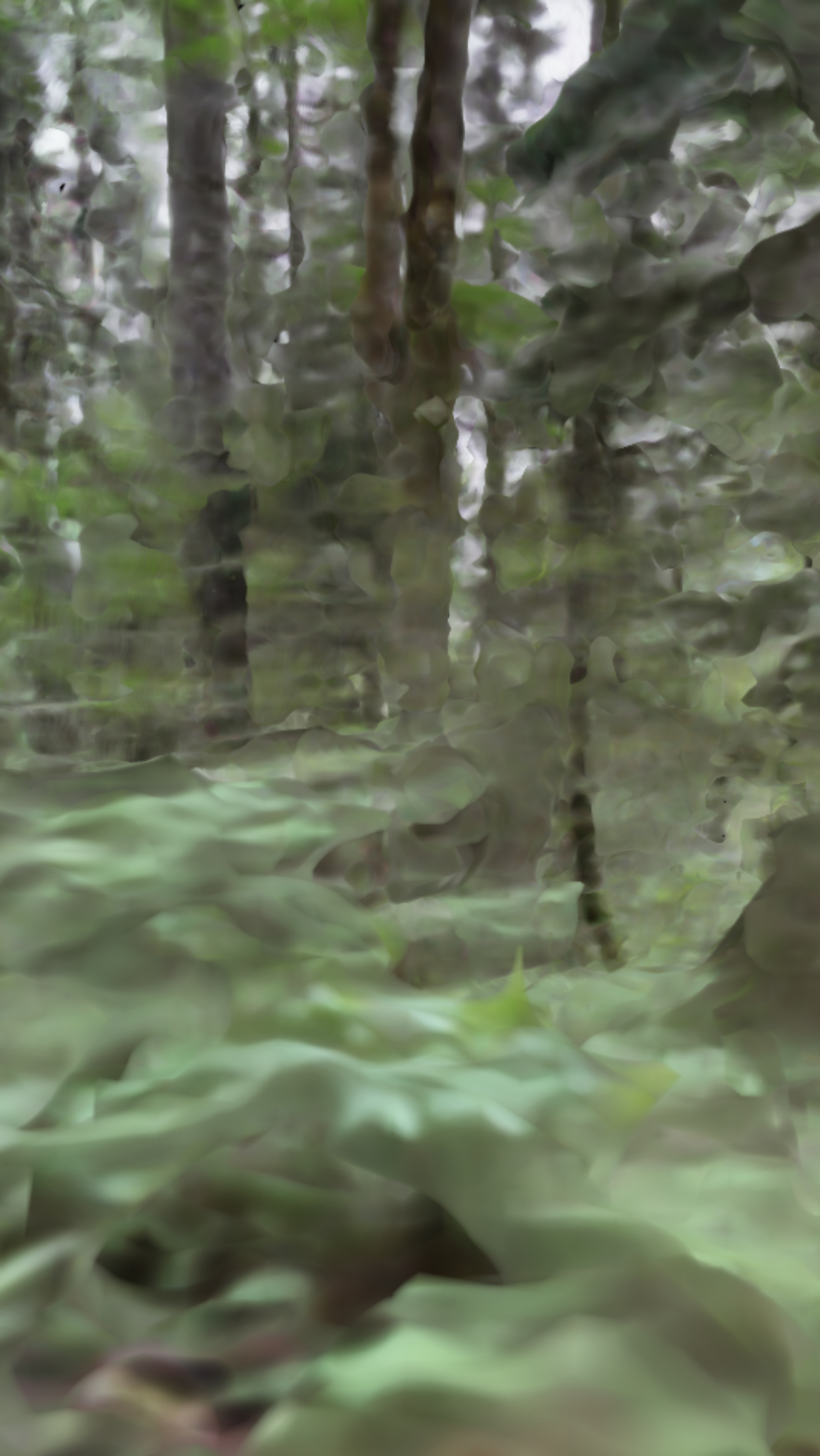}\\%
   RoDynRF}%
\hfill%
\parbox[t]{\figwidthNovel}{\centering%
  \includegraphics[trim=0 0 0 0, clip=true, width=\figwidthNovel]{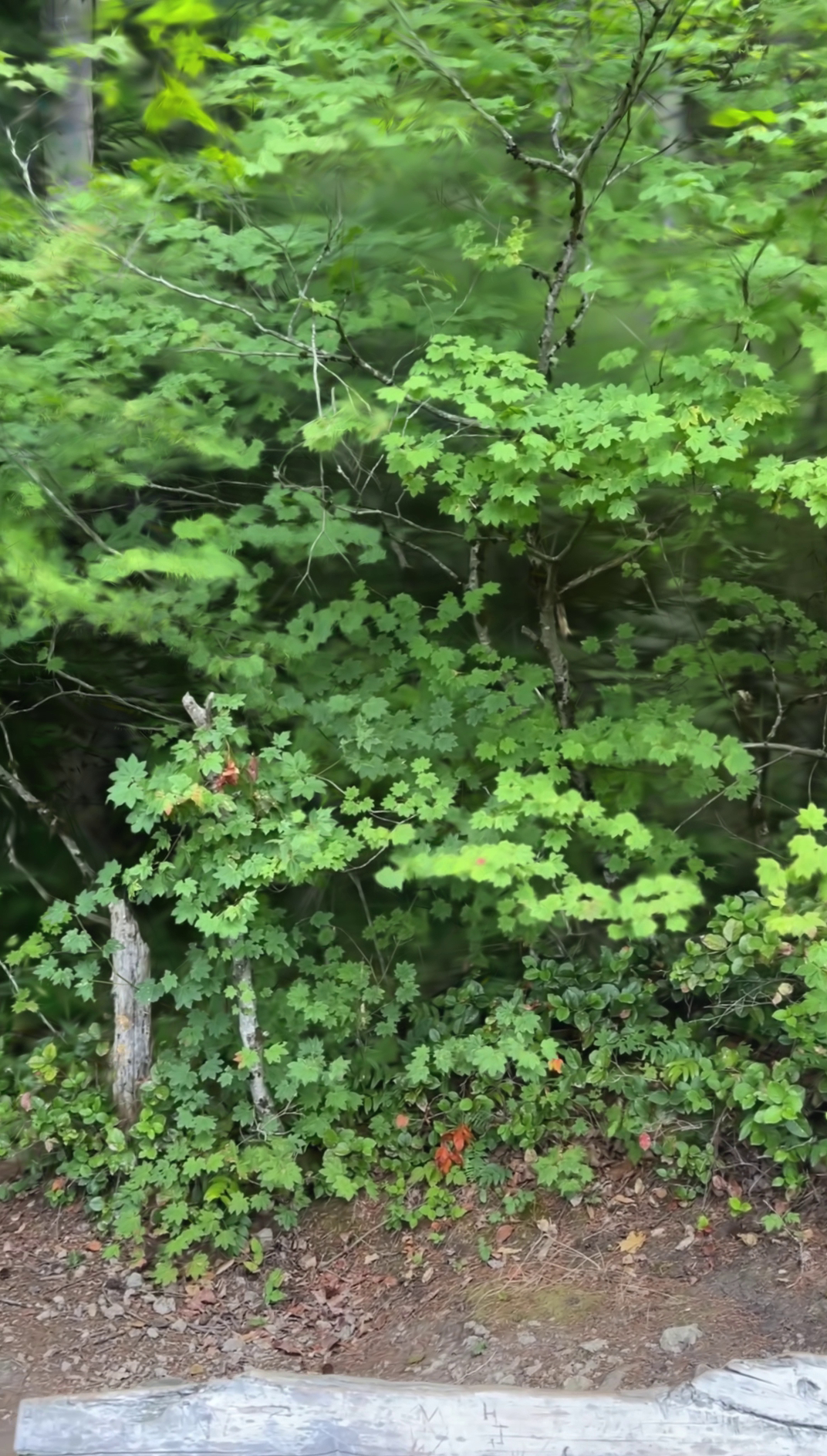}\\%
  \includegraphics[trim=0 0 0 0, clip=true, width=\figwidthNovel]{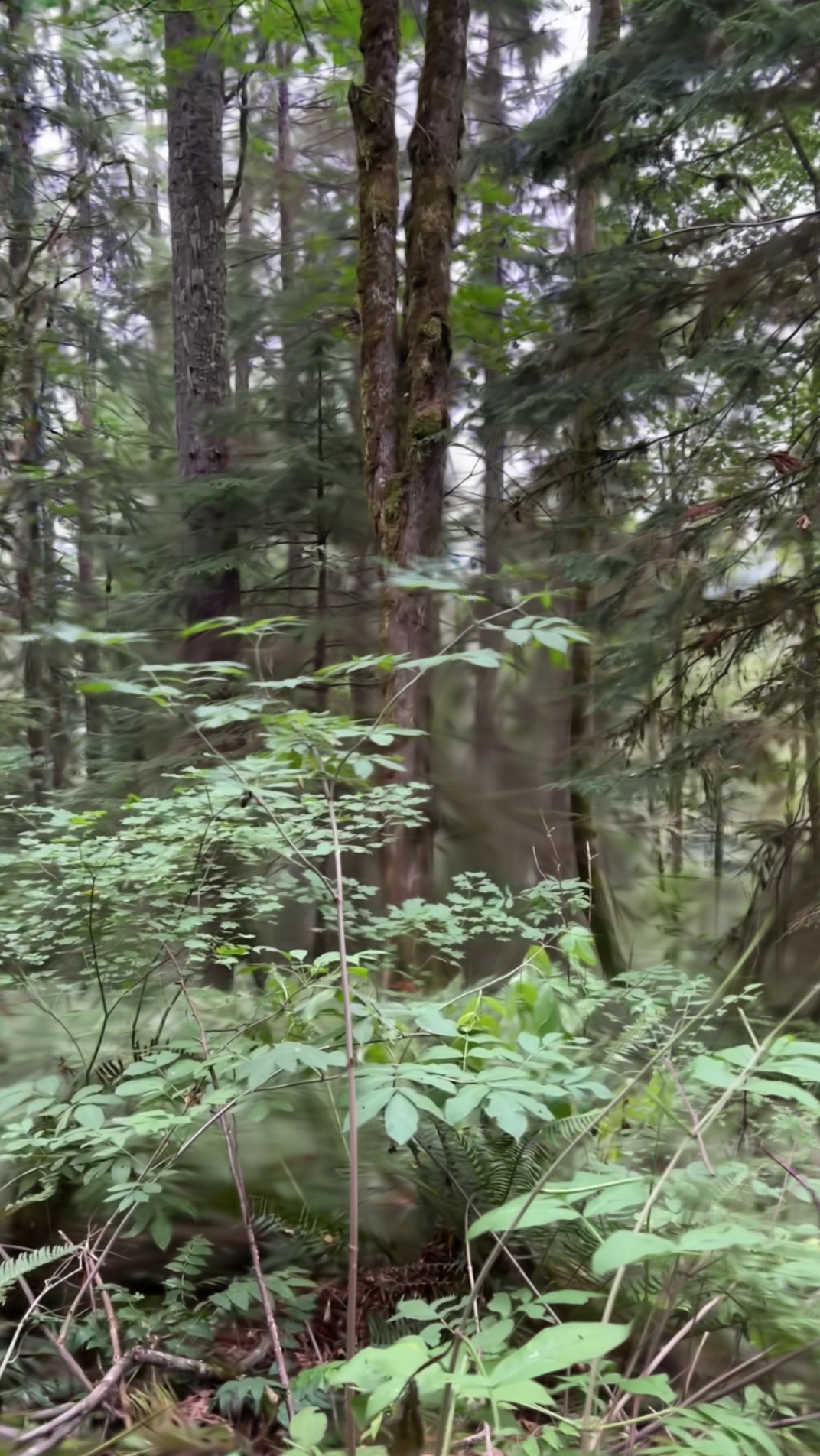}\\%
   4D-GS}%
\hfill%
\parbox[t]{\figwidthNovel}{\centering%
  \includegraphics[trim=0 0 0 0, clip=true, width=\figwidthNovel]{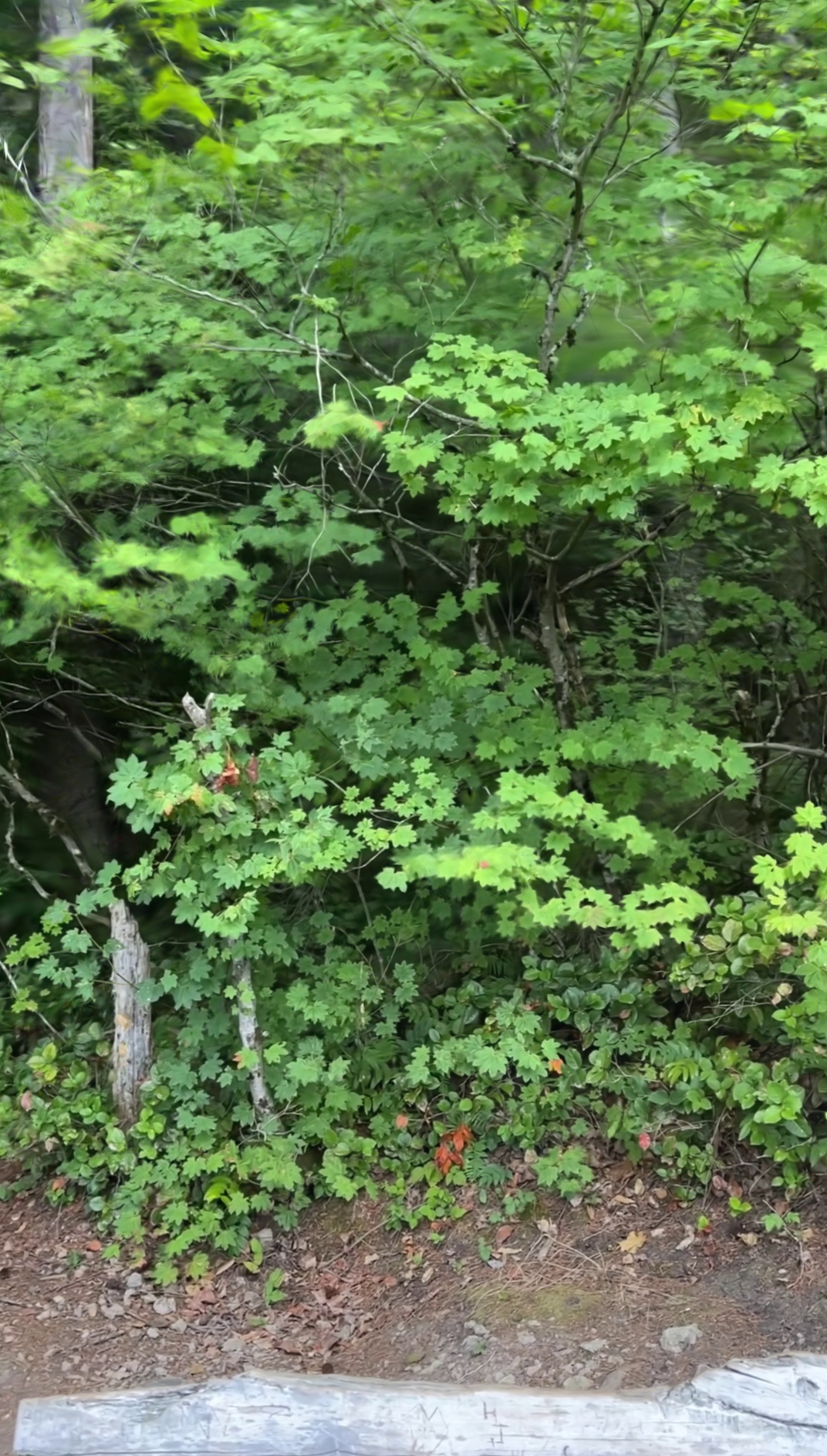}\\%
  \includegraphics[trim=0 0 0 0, clip=true, width=\figwidthNovel]{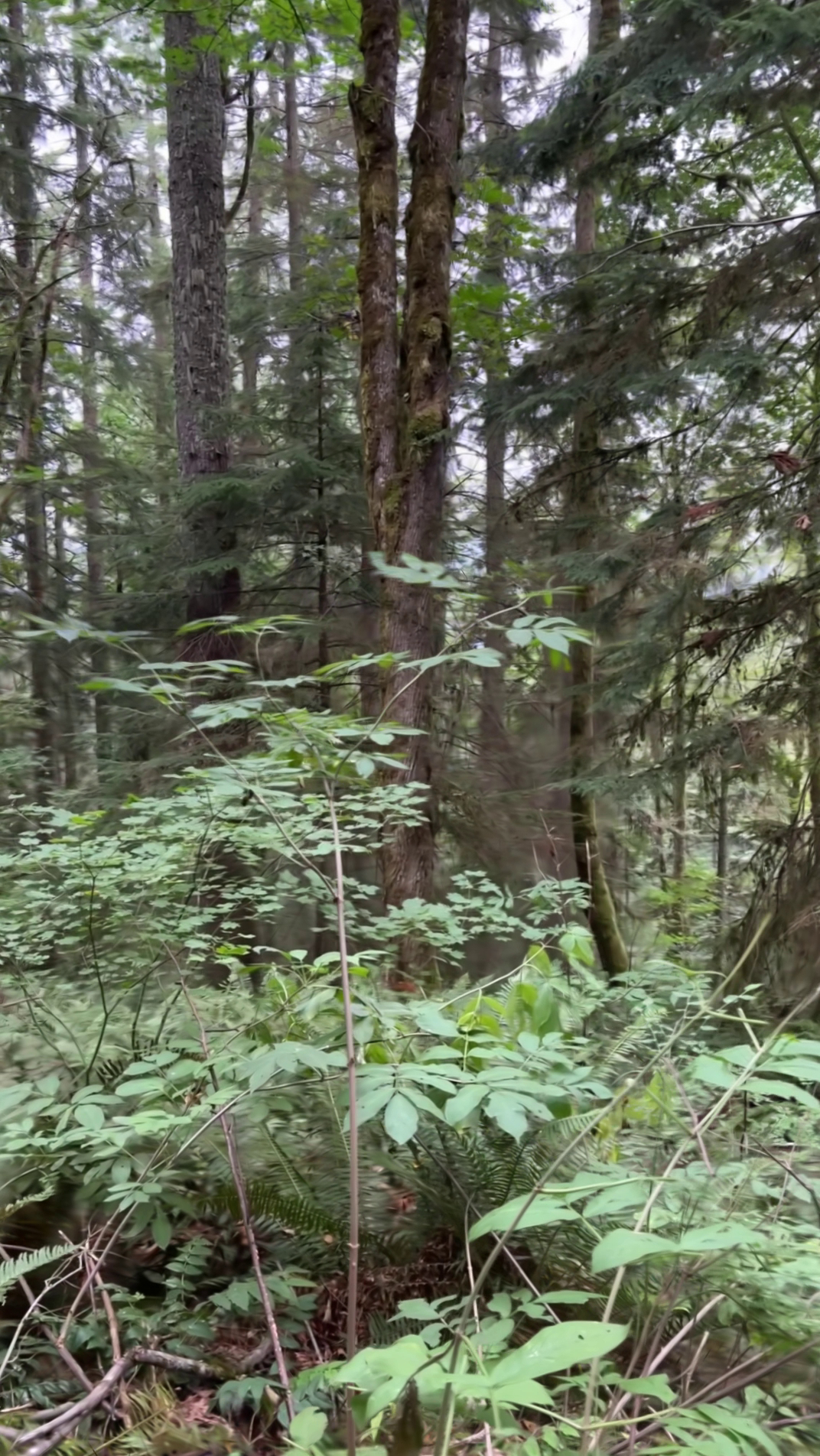}\\%
   Ours}%
\hfill%
\vspace{-3mm}
\captionof{figure}{
\textbf{Novel view synthesis comparisons.} 
We compare our method with RoDyNeRF~\cite{liu2023robust} and 4D-GS~\cite{wu20234dgaussians}.
Both show promising results for transient motion like moving people but struggle with complex ambient motion. 
RoDyNeRF struggles to produce sharp images due to the limitations of its implicit representation and constraints on optical flow.
4D-GS models part of the static scene but introduces artifacts in distant areas and fails to accurately represent ambient motion.
In contrast, our method successfully captures detailed appearances and handles ambient motion efficiently.
}
\Description{}
\label{fig:qualitative}
\end{figure*}

We show visual comparisons of our results against RoDyNeRF~\cite{liu2023robust} and 4D-GS~\cite{wu20234dgaussians} in Figs.~\ref{fig:qualitative} and ~\ref{fig:motivation}.
RoDyNeRF struggles to produce sharp images due to the limitations of its implicit representation and constraints on optical flow.
While 4D-GS manages to capture parts of the static scene effectively, it introduces significant artifacts in distant areas and fails to accurately represent ambient motion.
Our method successfully captures detailed appearances and handles ambient motion efficiently. The improvements our approach offers over 4D-GS stem from various design strategies discussed in Sec~\ref{sec:method}. To further quantify how our method compares with these baselines, we have conducted a user study on the dataset, detailed in Sec.~\ref{sec:quanitative}.

\subsection{Quantitative Comparison}
\label{sec:quanitative}
\begin{figure}[H]
    \centering
    \parbox[t]{\linewidth}{\centering%
  \includegraphics[trim=0 0 0 0, clip=true, width=\linewidth]{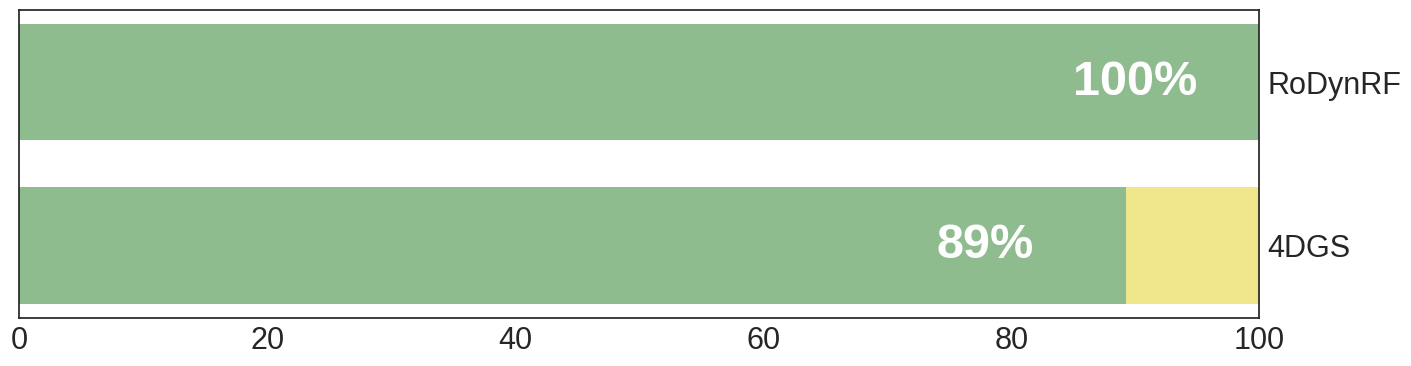}\\%
   (a) Fixed view, varying time} \\
    \parbox[t]{\linewidth}{\centering%
  \includegraphics[trim=0 0 0 0, clip=true, width=\linewidth]{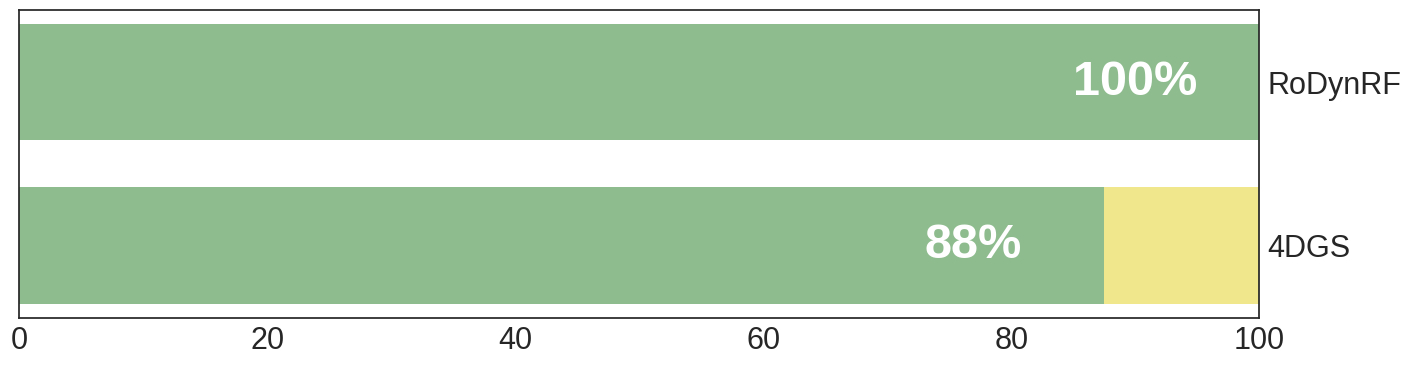}\\%
   (b) Fixed time, varying view} \\
    \parbox[t]{\linewidth}{\centering%
  \includegraphics[trim=0 0 0 0, clip=true, width=\linewidth]{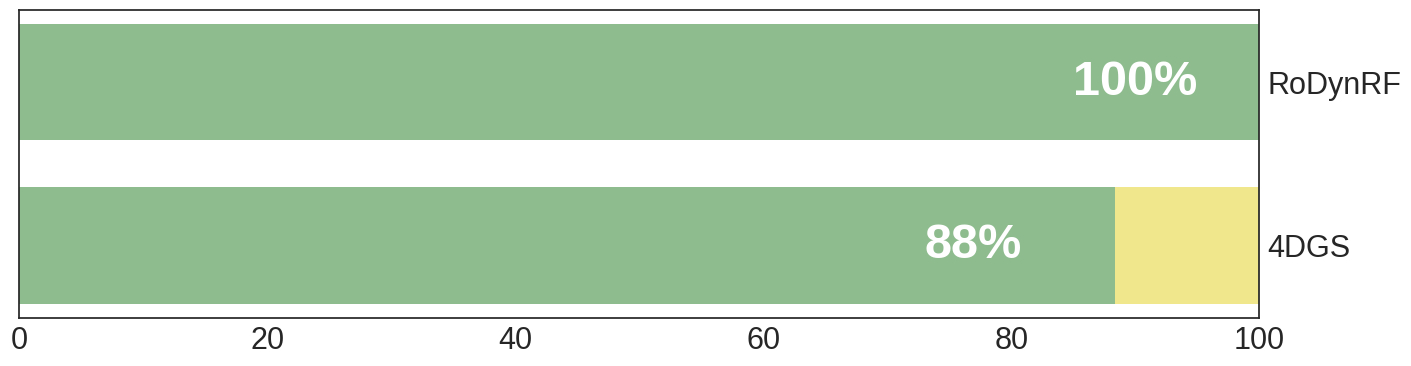}\\%
   (c) All samples}
\caption{
\textbf{User study}. 
Winrate of our method in comparision to other methods (labeled on the Y-axis). 
Lengths of the bars indicate the percentage of times users rated visual quality of a method highter than the competing method. The green bars with percentages correspond to our method.}
\Description{}
\label{fig:user-study}
\end{figure}

\paragraph{User study.} We randomly selected 7 examples from our dataset for the user study. For each example, we rendered two types of novel view synthesis results. Given a video with a duration of $T$ and a total of $N$ frames, we first fixed the time at $\frac{T}{2}$ and rendered a video using the camera trajectory from the input video. This first video assesses the quality of the reconstructed geometry and appearance. Secondly, we used the viewpoint from the $\frac{N}{2}$-th frame of the input video and rendered a video that progresses from time 0 to $T$. This second video evaluates the quality of the reconstructed motion.
With a side-by-side comparison of two baselines, we presented a total of 14 videos for users to judge in each of the two settings. Based on feedback from 24 users, we display the win rate in Fig.~\ref{fig:user-study}. It is clear from these results that our method significantly outperforms the others.

\paragraph{Full-reference(FR) metrics.}
To evaluate the performance in terms of FR metrics (i.e. PSNR/SSIM/LPIPS~\cite{zhang2018perceptual}), we hold out 3 segments, each containing 30 frames (totaling 90 frames), from the video to form the test-val set, using the remaining frames as the training set. We show the results in Tab.~\ref{tab:quantitative}, where it is evident that our method outperforms the two baselines.


\subsection{More diverse scenes}
\begin{figure}[H]
\setlength\figwidthLimitation{0.32\linewidth}
\centering%
\hfill
\parbox[t]{\figwidthLimitation}{\centering%
  \includegraphics[trim=0 0 0 0, clip=true, width=\figwidthLimitation]{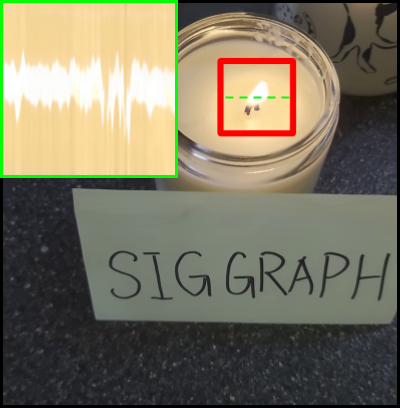}\\%
   Candle}%
\hfill%
\parbox[t]{\figwidthLimitation}{\centering%
  \includegraphics[trim=0 0 0 0, clip=true, width=\figwidthLimitation]{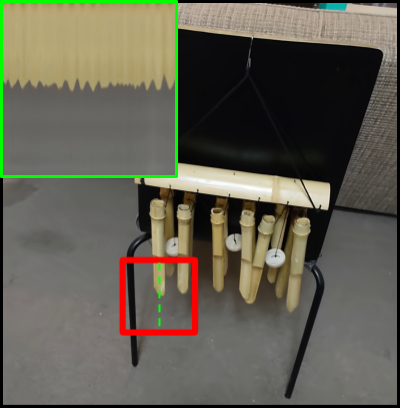}\\%
   Chimes}%
\hfill%
\parbox[t]{\figwidthLimitation}{\centering%
  \includegraphics[trim=0 0 0 0, clip=true, width=\figwidthLimitation]{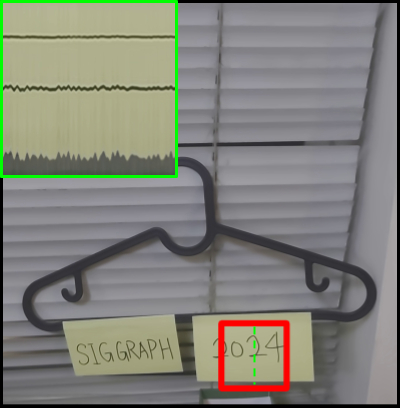}\\%
   Notes}%
\hfill%
\vspace{-3mm}
\captionof{figure}{
\textbf{Diverse scenes.} 
We showcase our method's ability to model candles, chimes, and notes using XT- or YT-slices.
}
\Description{}
\label{fig:more_examples}
\end{figure}

Our method can handle more than just plant motion. We demonstrate our method on a variety of examples in Fig.~\ref{fig:more_examples}, including the dancing flames of a candle, chimes swaying in the wind, and fluttering notes.
We also show the XT-YT slices of the rendered videos. Our method produces realistic, smooth, and periodic motion.
For a more detailed visualization of the motion, please visit our project website.


\subsection{Ablation Study}
\label{sec:ablation}
We conduct a quantitative analysis to highlight the importance of each component in our method, as shown in Table~\ref{tab:ablation}.
Specifically, we ablate depth regularization, DCT trajectory, trajectory MLP, rigidity regularization, and scene normalization, following the settings described in Section~\ref{sec:quanitative} to demonstrate the effectiveness of each component.

\begin{figure}[H]
\setlength\figwidthLimitation{0.49\linewidth}
\centering%
\hfill
\parbox[t]{\figwidthLimitation}{\centering%
  \includegraphics[trim=0 0 0 200, clip=true, width=\figwidthLimitation]{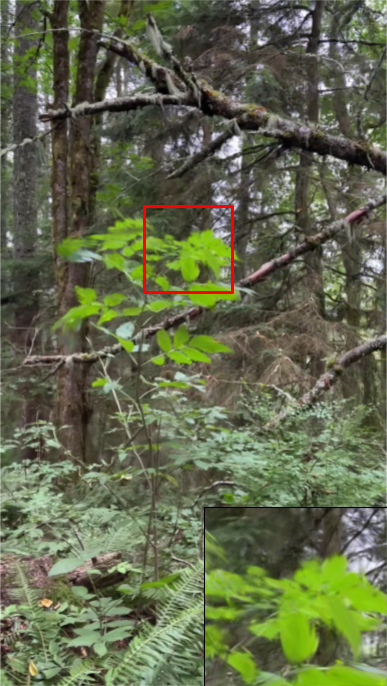}\\%
   w/o DCT transform}%
\hfill%
\parbox[t]{\figwidthLimitation}{\centering%
  \includegraphics[trim=0 0 0 200, clip=true, width=\figwidthLimitation]{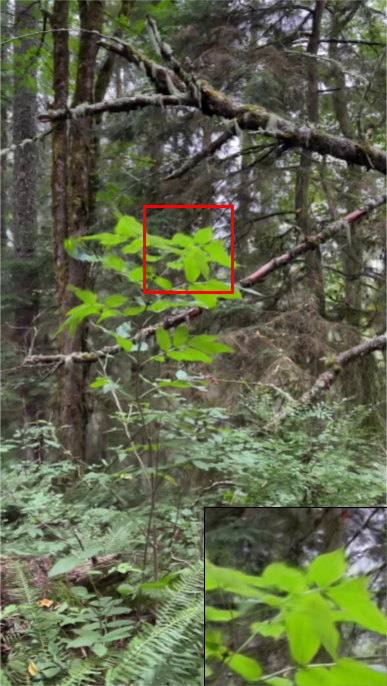}\\%
   w/ DCT transform}%
\hfill%
\captionof{figure}{
\textbf{Effect of DCT coefficients.} 
We synthesize an image of concealed leaves from a specific time when they're not visible in training data, comparing two MLP predictions: DCT coefficients and direct time-varying parameters. Direct predictions lack constraints on invisible motion, causing blurry or erroneous results. Using DCT coefficients, the model generalizes leaf motion from times when leaves are visible.
}
\Description{}
\label{fig:dct_transform}
\end{figure}

\begin{figure}[H]
\setlength\figwidthLimitation{0.49\linewidth}
\centering%
\hfill
\parbox[t]{\figwidthLimitation}{\centering%
  \includegraphics[trim=0 50 0 0, clip=true, width=\figwidthLimitation]{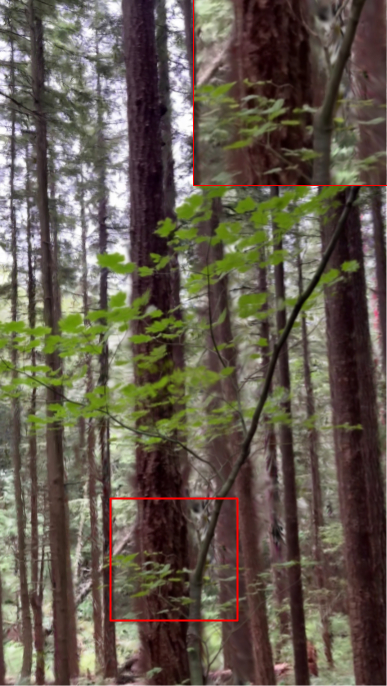}\\%
   w/o Rigidity and Relative Rotation Loss}%
\hfill%
\parbox[t]{\figwidthLimitation}{\centering%
  \includegraphics[trim=0 50 0 0, clip=true, width=\figwidthLimitation]{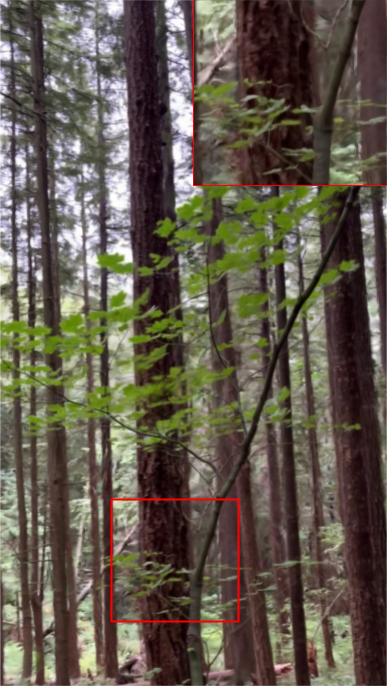}\\%
   w/ Rigidity and Relative Rotation Loss}%
\hfill%
\captionof{figure}{
\textbf{Effect of rigidity and relative rotation loss.} 
Introducing rigidity and relative rotation loss methods~\cite{luiten2023dynamic} ensures that neighboring Gaussians maintain consistent behavior, reducing artifacts from random rotations. This regularization helps the Gaussians move together, creating a more natural scene with fewer odd rotations.
}
\Description{}
\label{fig:relrigid}
\end{figure}

We illustrate the effects of depth regularization and rigidity regularization in Fig.\ref{fig:nerf_depth} and Fig.\ref{fig:relrigid}.
For the DCT trajectory prediction, we explore an alternative by replacing it with the direct prediction of $\Delta x_t$ and $\Delta q_t$ as proposed in 4D-GS (Fig.~\ref{fig:dct_transform}).
Additionally, we ablate the trajectory MLP by optimizing DCT coefficients directly per Gaussian, which we discovered leads to motion flickering.
Furthermore, we replace the normalization method for the Gaussian position $\mu$ fed into the MLP with the one proposed in 4D-GS (Fig.~\ref{fig:scene_contraction}).
\begin{table}[t]
    \caption{
    \textbf{Quantitative Evaluation using FR metrics.}
    Comparison to baselines on the forest sequence. 
    }
    \label{tab:quantitative}
    \centering
    \resizebox{0.6\linewidth}{!} 
    {
    \begin{tabular}{l | ccc}
    \toprule
    & PSNR $\uparrow$ & SSIM $\uparrow$ & LPIPS $\downarrow$ \\
    \midrule
    4D-GS                          &  19.04     &   0.472     &   0.414    \\
    RoDyNeRF                       &  18.12     &   0.393     &   0.512    \\
    Ours                           & \textbf{21.94} & \textbf{0.675} & \textbf{0.278} \\
    \bottomrule
    \end{tabular}
    }
    \vspace{-3mm}
\end{table}

\begin{table}[t]
    \caption{
    \textbf{Ablation study.}
    We report PSNR, SSIM and LPIPS on the forest sequence. 
    }
    \vspace{-0mm}
    \label{tab:ablation}
    \centering
    \resizebox{0.8\linewidth}{!} 
    {
    \begin{tabular}{l | ccc}
    \toprule
    & PSNR $\uparrow$ & SSIM $\uparrow$ & LPIPS $\downarrow$ \\
    \midrule
    Ours w/o depth regularization                          &  20.17     &   0.588     &   0.335    \\
    Ours w/o DCT transform                           &  21.65     &   0.663     &   0.282    \\
    Ours w/o trajectory MLP                           &  18.27     &   0.423     &   0.452    \\
    Ours w/o rigidity regularization         &  21.74     &   0.666     &   0.280   \\
    Ours w/o scene normalization             &  21.67     &   0.662     &   0.281    \\
    Ours (full model)                                   & \textbf{21.94} & \textbf{0.675} & \textbf{0.278} \\
    \bottomrule
    \end{tabular}
    }
    \vspace{-0mm}
\end{table}

\subsection{Limitations and Future Work}
\label{sec:limitations}
\begin{figure}[H]
\setlength\figwidthLimitation{0.32\linewidth}
\centering%
\hfill
\parbox[t]{\figwidthLimitation}{\centering%
  \includegraphics[trim=0 0 0 150, clip=true, width=\figwidthLimitation]{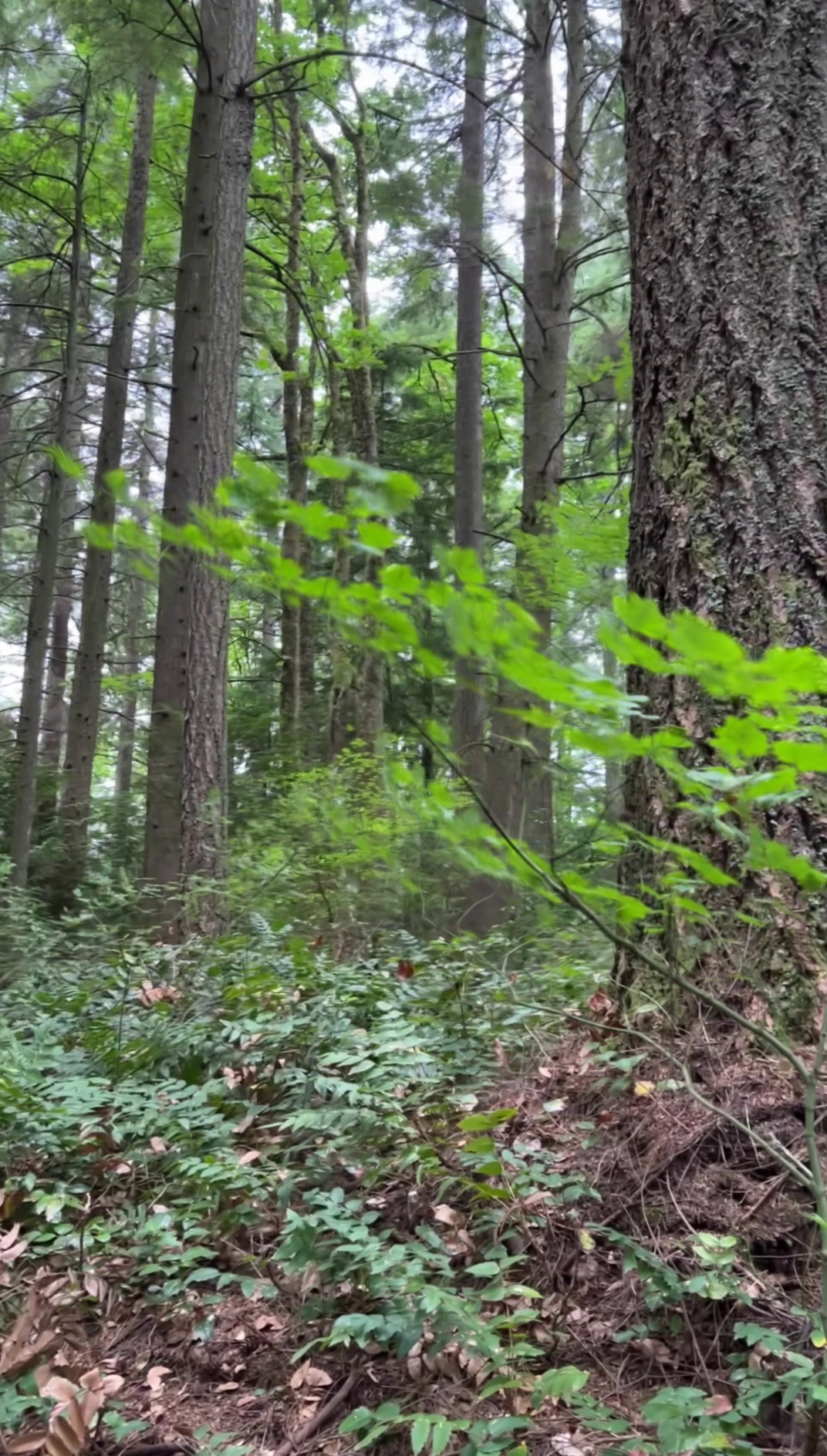}\\%
   Recontruction}%
\hfill%
\parbox[t]{\figwidthLimitation}{\centering%
  \includegraphics[trim=0 0 0 150, clip=true, width=\figwidthLimitation]{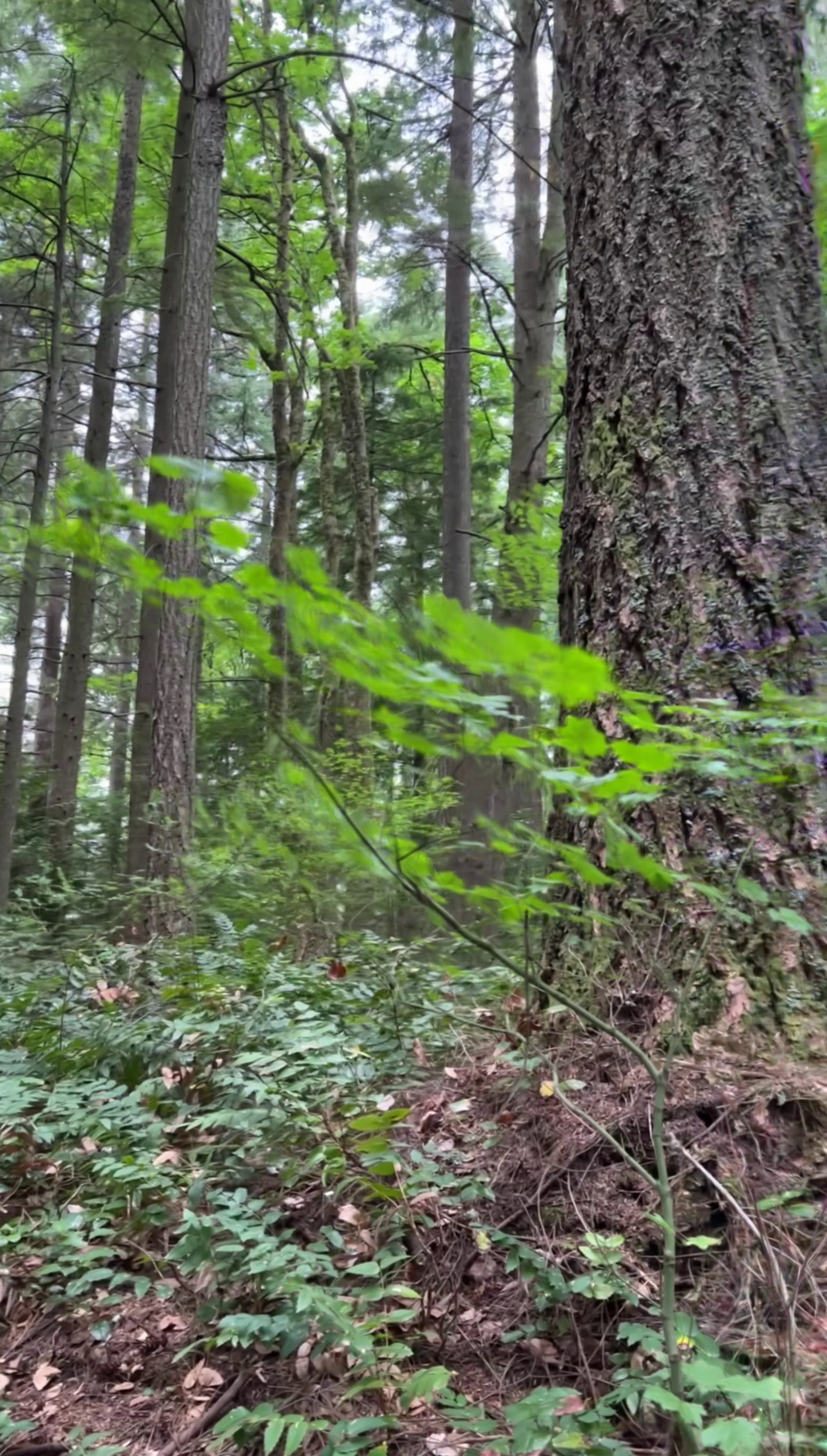}\\%
   Novel view}%
\hfill%
\parbox[t]{\figwidthLimitation}{\centering%
  \includegraphics[trim=0 0 0 150, clip=true, width=\figwidthLimitation]{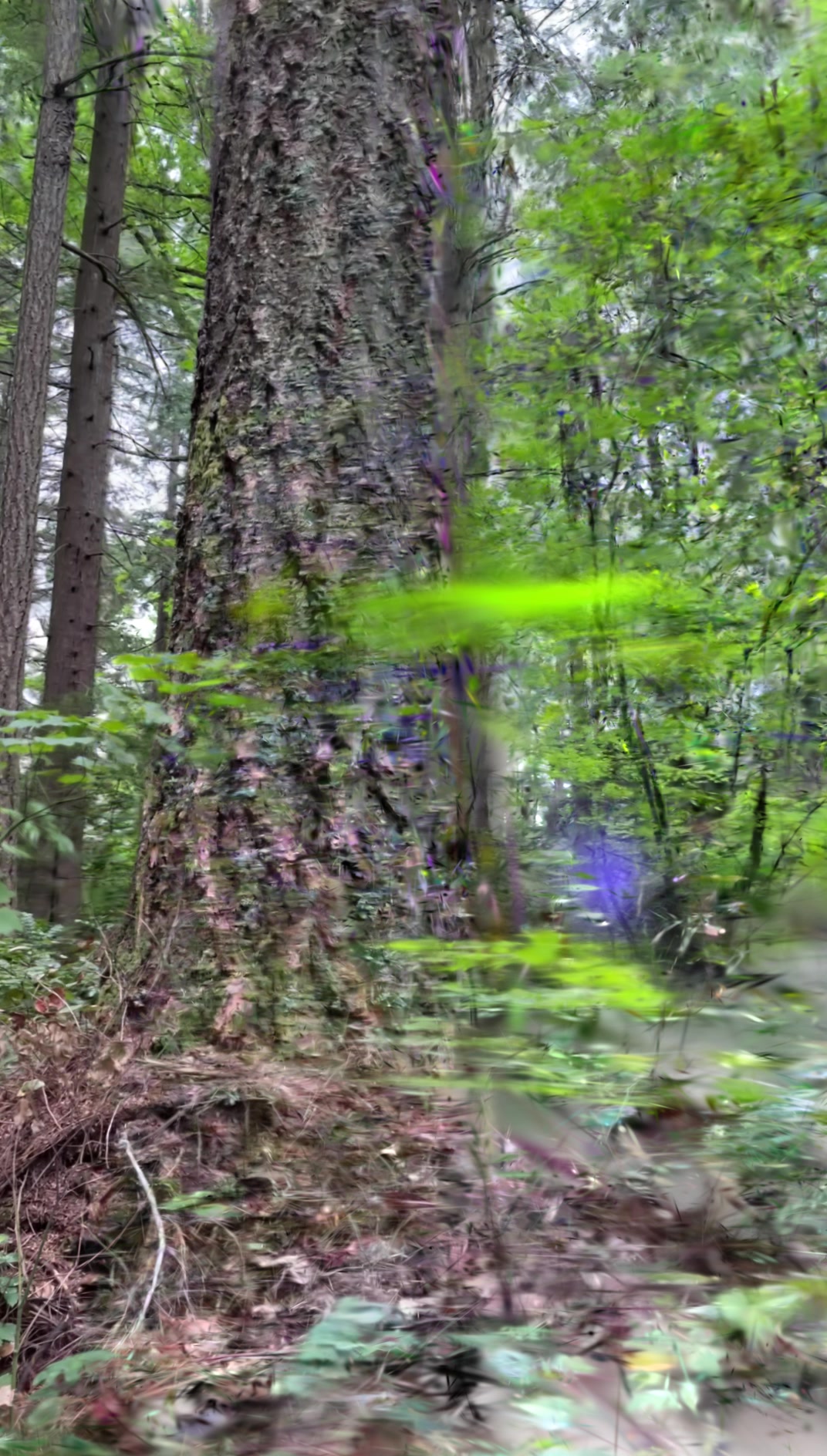}\\%
   Extreme view}%
\hfill%
\vspace{-3mm}
\captionof{figure}{
\textbf{Failure case of extreme view synthesis (view extrapolation).}
We observe color shift due to the extreme, unregularized viewpoints, unseen during training.
}
\Description{}
\label{fig:failure_case}
\end{figure}

Our method has certain limitations.
One constraint is the inability to handle complex non-periodic motion.
Additionally, while we can synthesize views freely at any time step within the range of input viewpoints, the quality of view synthesis deteriorates when attempting to extrapolate beyond this range (see Fig.~\ref{fig:failure_case}). Therefore, exploring the potential of generative models in future research could provide a promising avenue to enhance our method's ability to generalize to these complex scenarios.

\bibliographystyle{ACM-Reference-Format}
\bibliography{main}
\clearpage





 
\end{document}